\documentclass[twocolumn]{IEEEtran}
\usepackage{cite}
\usepackage{amsmath,amssymb,amsfonts}
\usepackage{algorithmic}
\usepackage{graphicx}
\usepackage{textcomp}
\def\BibTeX{{\rm B\kern-.05em{\sc i\kern-.025em b}\kern-.08em
    T\kern-.1667em\lower.7ex\hbox{E}\kern-.125emX}}
\usepackage[utf8]{inputenc}
\usepackage[T1]{fontenc}

\DeclareGraphicsExtensions{.png,.pdf}
\usepackage{subcaption}
\usepackage{dcolumn}
\usepackage{bm}
\usepackage{xcolor}
\usepackage{balance}
\usepackage{booktabs}
\usepackage{multirow}
\usepackage[ruled]{algorithm2e}
\usepackage{ragged2e}
\usepackage[normalem]{ulem}

\begin{document}

\title{Modified Feature Selection for Improved Classification of Resting-State Raw EEG Signals in Chronic Knee Pain}
\author{{Jean~Li}, 
{Dirk De Ridder}, 
{Divya Adhia},
{Matthew Hall}, %
{Jeremiah~D.~Deng}*~\IEEEmembership{Senior Member,~IEEE}
\thanks{J Li is with the School of Computing, University of Otago, Dunedin 9054, New Zealand; 
D De Ridder, D Adhia and M Hall are with the Department of Surgical Sciences, School of Medicine, University of Otago, Dunedin 9054, New Zealand. *JD Deng is with the School of Computing, University of Otago (correspondence e-mail: jeremiah.deng@otago.ac.nz).}}%


\IEEEpeerreviewmaketitle
\IEEEtitleabstractindextext{
\begin{justify}
\begin{abstract}
~\textit{Objective:} Diagnosing pain in research and clinical practices still relies on self-report. This study aims to develop an automatic approach that works on resting-state raw EEG data for chronic knee pain prediction. \textit{Method:} A new feature selection algorithm called ``modified Sequential Floating Forward Selection'' (mSFFS) is proposed. The improved feature selection scheme can better avoid local minima and explore alternative search routes. \textit{Results:} The feature selection obtained by mSFFS displays better class separability as indicated by the Bhattacharyya distance measures and better visualization results. It also outperforms selections generated by other benchmark methods, boosting the test accuracy to 97.5\%. \textit{Conclusion:} The improved feature selection searches out a compact, effective subset of connectivity features that produces competitive performance on chronic knee pain prediction. \textit{Significance:} We have shown that an automatic approach can be employed to find a compact connectivity feature set that effectively predicts chronic knee pain from EEG. It may shed light on the research of chronic pains and lead to future clinical solutions for diagnosis and treatment. 

\end{abstract}
\end{justify}
}

\maketitle
\IEEEdisplaynontitleabstractindextext
\IEEEpeerreviewmaketitle

\section{Introduction} 

Pain is defined as an unpleasant sensory and emotional experience associated with actual or potential tissue damage or described in terms of such damage~\cite{Bonica}. 
While acute pain can be recognized as a symptom of an underlying problem, chronic pain, which extends beyond the period of healing from the original insult or injury, lacks the acute warning function of physiological nociception~\cite{Treede2019-gu}. The International Association for the Study of Pain and International Classification of Diseases (ICD11) has defined chronic pain as pain that extends beyond three months, irrespective of the cause. Therefore, chronic pain can be considered not a mere symptom of another disease, but a health condition by itself~\cite{Scholz2019-tw,Treede2019-gu}. Indeed, chronic pain is not simply a temporal extension of acute pain but involves distinct mechanisms~\cite{Kuner2016-wx}. It is associated with multiple other symptoms. Despite being associated with physical conditions such as cardiovascular disease~\cite{Mills2019-kr}, chronic pain also correlates with cognitive dysfunction such as difficulties in attention, learning, memory, and decision making~\cite{Moriarty2011-id}. About 1/3 of the patients who suffer from chronic pain display other symptoms such as a combination of irritability, depression, anxiety, and sleep problems~\cite{Breivik2006-ya,Mills2019-kr,Yongjun2020-zr}. 

Chronic pain and its co-morbidities lead to increased disability~\cite{Mutubuki2020-iu}, carrying the highest global burden associated with disability~\cite{GBD_2017}. The correlation between chronic pain and disability creates a huge cost to society (\$560-635 billion per year), far more than heart disease, cancer, and diabetes~\cite{Gaskin2012-wg}, and back pain alone amounts to 1.7\% of the global national product every year~\cite{Van_Tulder1995-fk}. 
Chronic pain is a worldwide problem. Studies have found that the prevalence of chronic pain is 20\%, both in the USA~\cite{Dahlhamer2018-vn} and Europe~\cite{Breivik2006-ya}, and around 30\% in China~\cite{Yongjun2020-zr}, which is similar to the prevalence in low-income and middle-income countries~\cite{Jackson2015-dm}. High-impact chronic pain, which occurs in 8\% of the population, frequently limits patients' life or work activities~\cite{Dahlhamer2018-vn}. 
Due to the lack of effective, specific, and safe therapies, chronic pain remains one of the major causes of human suffering~\cite{GBD_2017}. Many of the currently available pain therapies either are inadequate or cause unpleasantness to patients~\cite{Wager2013-nk}, some even causing deleterious side effects~\cite{Stucky2001-jg,Jensen2014-uy}. 

In research and clinical practice, the gold standard of diagnosing pain is based on self-report. Self-report pain diagnosis could be challenging in certain clinical contexts or conditions. In particular, reliance on self-report of pain can be challenging where there is no evidence of tissue damage, or the presence of tissue damage may not be necessary to determine the presence of pain. Moreover, an objective measure of pain could pave the way to improve diagnostic and treatment outcomes. Therefore, it is necessary to develop simple, reliable, and clinically applicable method that can validate the pain experience and justify the allocation of relevant healthcare services and benefits. 

The goal of this study is to find an automatic method to predict chronic knee pain through resting-state EEG. We focus on functional connectivity rather than activity, as recent studies have suggested the effectiveness of functional connectivity in EEG pain analysis~\cite{kim2021sex,kim2020cross,kim2021neural}, and this may be superior to detecting a pain-generating network~\cite{De_Ridder2017-gv}. 

Our main contribution is to propose a modified sequential floating forward feature selection algorithm (mSFFS) that explores alternative search paths to avoid potential local minima and searches out better feature selection outcome. Our experiments show that mSFFS outperforms a wide range of existing feature selection solutions. Apart from classification performance evaluation, we also assess the quality of the feature schemes from multiple perspectives, including visualization, and evaluation of the Jaccard index and a separability index based on the Bhattacharyya distance. 

The remainder of the paper is organized as follows. We first give a brief review of some related work in Section II, and introduce our EEG dataset in Section III. Methods for pre-processing of EEG recordings and feature extraction are outlined in Section IV, where our modified algorithm for feature selection, the approach for model evaluation, and the class separability index are also given.  Section V presents the experiment results, where mSFFS show clear competence. A 20-dimension connectivity feature set is found, and its relevance to the recent literature is discussed. We conclude the paper in Section VI. 


\section{Related work}

Multiple previous studies have shown the effectiveness of pain prediction through EEG using machine learning methods.
Vanneste \textit{et al.}~\cite{Vanneste2018} has yielded an accuracy of 92.53\% on chronic pain through a support vector machine (SVM) trained on regions of interest.
Nezam \textit{et al.}~\cite{nezam2018novel} applied a SVM on selected band power features and reported an accuracy of 83$\pm$5\% and 62$\pm$6\% for the three and five pain levels respectively. Vuckovic \textit{et al.}~\cite{vuckovic2018prediction} predicted central neuropathic pain based on the resting-state EEG and reported an accuracy of 85\% with a simple linear classifier using band power features. Vijayakumar \textit{et al.}~\cite{vijayakumar2017quantifying} investigated quantifying tonic thermal pain across healthy subjects. By using time-frequency wavelet representations of independent components obtained from EEG, their random forest model achieved an accuracy of 89.45\%.  
Features selected from a combination of power spectral density (PSD) and functional connectivity (FC) abtained an accuracy of 86.2\%–93.8\% in classifying heat pain~\cite{hsiao2021machine}.
And 85\%-100\% accuracies can be achieved when classifying stimulated pain versus non-pain state on the same individuals using SVM~\cite{okolo2018use}.

These previous approaches, however, relied on labor-intensive pre-processing for artefact removal, which is expensive and, therefore, cannot be used in routine clinical practice. Therefore, a user-friendly, clinically applicable system is needed that can automatically pick up the neural pain signature on raw, unprocessed data, using a low-cost device.

\section{Data Description} 
The data used for this study consist of EEG recordings of 37 healthy participants, and 37 patients with chronic knee pain. Data were acquired in Dunedin Hospital with 19 electrodes in the standard 10–20 placement (Fp1, Fp2, F7, F3, Fz, F4, F8, T3, C3, Cz, C4, T4, T5, P3, Pz, P4, T6, O1, O2). 
The study was approved by the Health and Disability Ethics Committee (HDEC), New Zealand (21CEN63), on 22 March 2021.


\section{Methods} 
\subsection{Pre-processing and feature extraction}
All EEG recordings went through an automatic pre-processing procedure to remove potential artifacts and control volume conduction.

The first 10 seconds of all the recordings were discarded to avoid the ``initial drift'' noise. A notch filter at 50 Hz to attenuate the power line interference was applied to all data, followed by a high-pass filter of 1 Hz and a low-pass filter of 40 Hz.
Each recording was cut into 10-second long epochs and went through an auto-cleaning process to repair epochs by interpolation or excluding them. This is achieved by estimating an optimal global peak-to-peak threshold first, then finding trial-wise bad sensors by estimating a separate rejection threshold for each sensor~\cite{jas2017autoreject}. 
Lastly, the current source density transformation that is based on spherical spline surface Laplacian~\cite{perrin1987scalp}~\cite{perrin1989spherical}~\cite{cohen2014analyzing}~\cite{kayser2015benefits} was applied to control volume conduction.

In this study, the EEG connectivity is based on the corrected imaginary phase locking value (ciPLV)~\cite{bruna2018phase}, as this metric is robust in the presence of volume conduction, and proved capable of ignoring zero-lag connectivity while correctly estimating nonzero-lag connectivity~\cite{bruna2018phase}.
In particular, for each pair of channels, the ciPLV across all epochs on chosen frequency $\omega$ was computed based on the averages of the real and imaginary parts of the cross-channel spectrum densities: 
\begin{equation}\label{eq:coh}
    C(\omega)=\frac{\left| E\left[\frac{Im(S_{xy}(\omega))}{|S_{xy}(\omega)|} \right] \right|}{\sqrt{1- \left|E\left[ \frac{Re(S_{xy}(\omega)}{|S_{xy}(\omega)|} \right]\right| ^ 2 }},
\end{equation}
where $S_{xy}(.)$ is the cross-spectral density for channels $x$ and $y$, and $E[.]$ denotes averaging over all epochs within a sample.

For every pair of channels, the connectivity across all epochs was computed on the mean frequency value of 5 major brain wavebands.
These frequency bands are delta ($1\sim 4$ Hz), theta ($4\sim 8$ Hz), alpha ($8\sim 12$ Hz), beta ($12\sim 30$ Hz), and gamma ($30\sim 40$ Hz). 
Each sample is therefore represented by a feature vector of 855 dimensions (171 channel pairs $\times$ 5 bands).
The MNE package~\cite{GramfortEtAl2013a} was applied to compute the spectral connectivity measures.

\subsection{Feature selection}
There are two main challenges to selecting features for our study. First, we have a relatively small sample size of 74 and a large number of features at 855. This requires a feature selection approach that can effectively handle high-dimensional problems with limited data. Second, the signals collected from each channel represent a summation of neural activity from a broad region of the brain, resulting in high collinearity between features in sensory space EEG analysis.

Various existing feature selection schemes were examined, including those based on filtering, greedy search, mutual information, random shuffling and swarm intelligence. Unfortunately, none of these approaches yielded satisfactory results. As a result, we developed a customized feature selection scheme by modifying the sequential floating forward selection (SFFS) method~\cite{Pudil1994floating}. Like SFFS, we also use a classifier as a \textit{wrapper}. Although our feature method is independent of wrapper choices, we used a support vector machine (SVM) with radial basis function kernels as the classifier/wrapper. For the remainder of this article, we will refer to this modified SFFS scheme as ``mSFFS''.

mSFFS starts with an empty feature set.
In each searching iteration, a forward selection is conducted, followed by a backward selection.
In the forward selection step, assuming there are $k-1$ features in the current set, the feature that can increase the historical best score of $k$-feature sets is included to form a new set of $k$ features.
Similarly, in backward selection, features are excluded one by one if the reduced set results in a higher score than the historical best set.
This process is repeated until $k$ reaches the number of features required. 
Compared to the classical SFFS algorithm, mSFFS records the winning sets of all numbers of features and always starts forward selection from a historical best set, hence allowing for multiple searching branches. The winning sets are updated along the way of both the forward and backward selection process. Bookkeeping of all searched sets is carried out to speed up the searching process and avoid deadlocks. The feature selection algorithm, using an external classifier function to evaluate feature selections, is presented in Algorithm~\ref{algo:MSFFS}.

To narrow down the search space, a preselection procedure was employed. We calculated the SHAP values~\cite{Lundberg_nips17} using XGBoost~\cite{Chen:2016}, and only kept the features with a mean absolute SHAP value greater than a given threshold. We took a lenient setting here and had the threshold set to 0. Depending on the exact situation, a positive threshold value could be considered if a harsher preselection policy is intended. 

The number of selected features to be used in modeling will contribute to the complexity of the model, which may affect its generalization ability. We used cross-validation in deciding the optimal number of features because cross-validation is an established, reliable approach in model selection~\cite{James2021}. In our experiment, we obtained the performance of feature selections using 1 to 50 features incrementally, and selected the best feature set corresponding to the best cross-validation accuracy.

\SetKwProg{Presel}{Preselection:}{}{}
\SetKwProg{Func}{External Function:}{}{}
\SetKwProg{Init}{Initialization:}{}{}
\SetKwProg{ForwardSel}{Forward selection:}{}{}
\SetKwProg{BackwardSel}{Backward selection:}{}{}
\begin{algorithm}[t]
\caption{Modified SFFS}\label{algo:MSFFS}
\KwData{$K$: intended number of features; full feature set $\cal F$}
\Func{score(W)}{returns the performance metric (e.g., accuracy) of the classifier given a feature scheme $W$;}
\KwResult{Selected feature subset $W_K$}
\Presel{}{
  Calculate SHAP value for all features\;
  Remove features with negative SHAP values from $\mathcal{F}$\;
}

\Init{}{
    $k\leftarrow 1$ \;
    \For{$i\in$ 0:K}{winning set $W_i\leftarrow\emptyset$\; 
    winning set score $s_i\leftarrow 0$\;}
    Feature set bookkeeping: $\mathcal{B} \leftarrow\emptyset$\;
} 
\While{$k \leq K$}{ 
  \ForwardSel{}{
  Find the best feature to add: $f\leftarrow\arg\displaystyle\max_f~\mathrm{score}(W_{k-1}\cup f)$, $\forall f\in\mathcal{F},  (W_{k-1}\cup f)\notin \mathcal{B}$\;
  Update bookkeeping: $\mathcal{B}\gets \mathcal{B}\cup\{W_{k-1}\cup f\}$\;
  \If{$\mathrm{score}(W_{k-1}\cup f)>s_k$}{
    Update winning selection: $W_k \gets W_{k-1}\cup f$ \;
    Update winning score: $s_k\gets \mathrm{score}(W_{k})$ \;   
  }
  }
  \BackwardSel{}{
  \While{$k>2$}{
  Find the best feature to drop: $f \gets \arg\displaystyle\max_f \mathrm{score}(W_k\backslash f)$, 
             $\forall f \in W_k$, $(W_k\backslash f)\notin \mathcal{B}$\;
  Update bookkeeping: $\mathcal{B}\gets \mathcal{B}\cup \{W_k\backslash f\}$\;
  \eIf{$\mathrm{score}(W_{k}/f)>s_k$}{
    Update winning selection: $W_{k-1}\gets W_k\backslash f$\;
    Update winning score: $s_{k-1}\leftarrow \mathrm{score}(W_{k-1})$\;     
    $k\gets k-1$\;
  }{
  break\;
  }
  }
}
  $k\gets k+1$\;
}
\end{algorithm}

\subsection{Model training and evaluation}
\label{sec:mt}
Tentative feature selections were used to give classification performance. We performed nested cross-validation using support vector machien (SVM) classifiers to obtain the accuracy scores. 

In each of the outer 10-fold cross-validation loop, the data points of 90\% of the individuals were used for training, on which another 10-fold cross-validation followed for parameter tuning and cross-validation score generation. We then use the rest 10\% data to obtain the test scores.



\subsection{Feature selection evaluation}
To examine the efficiency of our proposed mSFFS feature selection algorithm, we also trained and obtained the test scores through the same process, using features selected by the classical SFFS (from the \texttt{mlxtend} package~\cite{raschkas_2018_mlxtend}) (denoted by SFFS), and the top-$n$ features ranked by SHAP using the same amount of features (denoted by SHAP). 

For a wider comparison, we also evaluated various swarm-based feature selection schemes, including the binary particle swarm optimization (BPSO)~\cite{Xue:13}, binary Harris Hawks optimizer (BHH)~\cite{thaher2020binary}, 
  binary grey wolf optimization (BGW)~\cite{emary2016binary}, and the improved dragonfly algorithm (IDA)~\cite{hammouri2020improved}. We used the open-sourced Zoofs library \footnote{Available at https://jaswinder9051998.github.io/zoofs/} with default setting to perform these swarm-based feature selections.
The classification performance of feature selections obtained from these algorithms was compared with that obtained by mSFFS. 
\begin{figure*}[!t]
\centering
\includegraphics[width=0.65\textwidth]{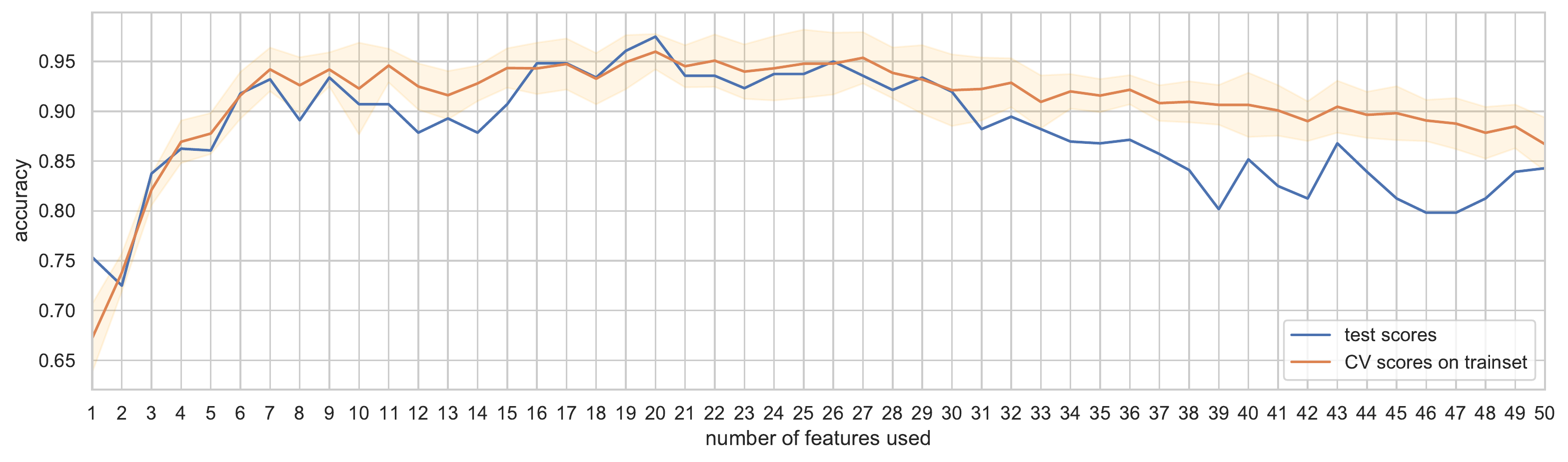}
\caption{Classification accuracy scores when using an increasing number of mSFFS selected features. The shadow area gives the confidence intervals of the cross-validation scores.} \label{fig:msffs_curve}
\end{figure*}
\begin{figure*}[!h]
\centering
\includegraphics[width=0.65\textwidth]{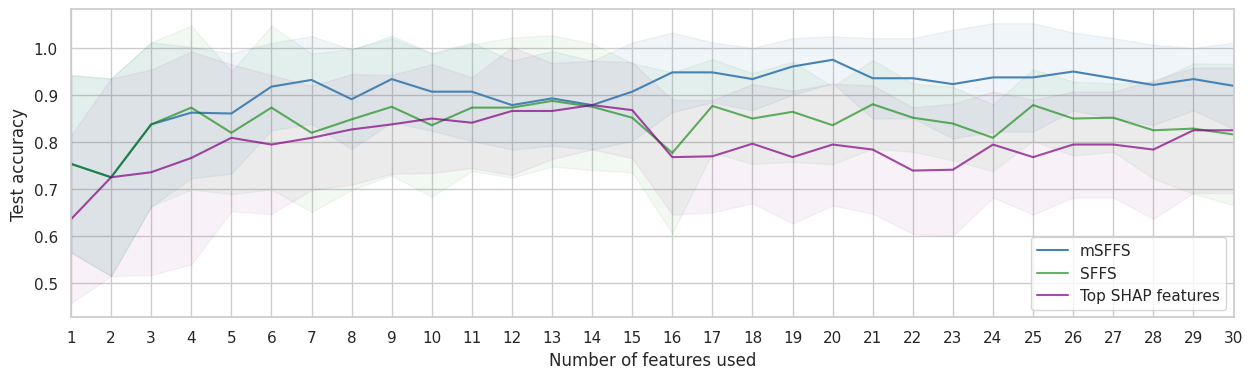}
\caption{Testing scores using an increasing number of features. The shadow areas represent the confidence intervals for 10-fold tests.} \label{fig:fs_compare}
\end{figure*}

Obviously, there will be partial overlaps between various feature selections as the relevant algorithms all target some good features. To evaluate the similarity between selected feature sets, we used the Jaccard index. Given two feature selections $F_1$ and $F_2$, the Jaccard index is given by
\begin{equation}
    \mathcal{J}(F_1, F_2)=\frac{|F_1 \cap F_2|}{|F_1 \cup F_2|}.
\end{equation}
We expect the Jaccard indices will be moderate between selected feature sets given by different selection algorithms. 

To further assess the quality of selected feature schemes, and investigate the differences between mSFFS and SFFS, we took both a qualitative approach (through visualization) and a quantitative approach (by measuring class separabilities). 

We choose the t-distributed stochastic neighbor embedding (t-SNE) algorithm~\cite{vandermaaten08} to visualize the data.
The Scikit-learn~\cite{scikit-learn} implementation of t-SNE is used in this study. The perplexity is set to 25. 

We use the Bhattacharyya distance as an additional metric to measure the separability of the feature selection schemes, independent from classifier choices.

Assume two normal distributions $p=N(\boldsymbol{\mu}_1,\boldsymbol{\Sigma}_1)$ and $q=N(\boldsymbol{\mu}_2,\boldsymbol{\Sigma}_2)$. The Bhattacharyya distance between them is given by
\begin{equation}
    D_B(p,q)=\frac{1}{8}(\boldsymbol{\mu}_1-\boldsymbol{\mu}_2)^T \boldsymbol{\Sigma}^{-1}(\boldsymbol{\mu}_1-\boldsymbol{\mu}_2)+\frac{1}{2}\ln\left(\frac{|\boldsymbol{\Sigma}|}{\sqrt{|\boldsymbol{\Sigma}_1||\boldsymbol{\Sigma}_2|}}\right),
\end{equation}
where $\boldsymbol{\Sigma}=(\boldsymbol{\Sigma}_1+\boldsymbol{\Sigma}_2)/2$.
From this definition, we can see that the Bhattacharyya distance extends the Mahalanobis distance by adding an extra term that informs on the dissimilarity between the two covariance matrices. 
The Bhattacharyya distance has been effectively used for optimizing feature extraction~\cite{Choi2003-xk} and transfer learning~\cite{Pandy22}. 

\begin{table*}[!t]
\scriptsize
    \centering
    \begin{tabular}
  {p{0.01\linewidth}||p{0.4\linewidth}|p{0.03\linewidth}||p{0.4\linewidth}|p{0.03\linewidth}}
\hline
\multirow{2}{*}{\textbf{$k$}} & \multicolumn{2}{c||}{\textbf{mSFFS}} & \multicolumn{2}{c}{\textbf{SFFS}} \\ \cline{2-5} 
 	 	&	\textbf{Selected features}	&	\textbf{t-score}	&	 \textbf{Selected features}	&	\textbf{t-score}	\\ \hline
 1  & P4-O1($\theta$)  & 0.754  & P4-O1($\theta$)  & 0.754 \\   \hline
 2  & P4-O1($\theta$), Fp1-F7($\theta$)  & 0.725  & Fp1-F7($\theta$), P4-O1($\theta$)  & 0.725 \\   \hline
 3  & P4-O1($\theta$), Fp1-F7($\theta$), T4-Pz($\theta$)  & 0.838  & Fp1-F7($\theta$), P4-O1($\theta$), T4-Pz($\theta$)  & 0.838 \\   \hline
 4  & P4-O1($\theta$), Fp1-F7($\theta$), T4-Pz($\theta$), F4-P4($\beta$)  & 0.862  & Fp1-F7($\theta$), P4-O1($\theta$), T4-Pz($\theta$), C3-T4($\beta$)  & 0.873 \\   \hline
 5  & P4-O1($\theta$), Fp1-F7($\theta$), T4-Pz($\theta$), F4-P4($\beta$), C3-T4($\beta$)  &  0.861  & Fp1-F7($\theta$), P4-O1($\theta$), T4-Pz($\theta$), Fp1-F4($\beta$), C3-T4($\beta$)  &  0.820 \\   \hline
 6  & P4-O1($\theta$), Fp1-F7($\theta$), T4-Pz($\theta$), F4-P4($\beta$), Fp2-T4($\theta$), Fz-C4($\delta$)  &  0.918  & Fp1-F7($\theta$), P4-O1($\theta$), T4-Pz($\theta$), Fp1-F4($\beta$), Cz-T6($\beta$), C3-T4($\beta$)  &  0.873 \\   \hline
 7  & P4-O1($\theta$), Fp1-F7($\theta$), T4-Pz($\theta$), F4-P4($\beta$), Fp2-T4($\theta$), Fz-C4($\delta$), F8-Cz($\delta$)  &  0.932  & Fp1-F7($\theta$), P4-O1($\theta$), T4-Pz($\theta$), Fp1-F4($\beta$), F4-C4($\delta$), Cz-T6($\beta$), C3-T4($\beta$)  &  0.82 \\   \hline
 8  & P4-O1($\theta$), Fp1-F7($\theta$), T4-Pz($\theta$), F4-P4($\beta$), Fp2-T4($\theta$), Fz-C4($\delta$), F8-Cz($\delta$), T4-Pz($\gamma$)  &  0.891  & Fp1-F7($\theta$), P4-O1($\theta$), T4-Pz($\theta$), F8-T3($\gamma$), Fp1-F4($\beta$), F4-C4($\delta$), Cz-T6($\beta$), C3-T4($\beta$)  &  0.848 \\   \hline
 9  & P4-O1($\theta$), Fp1-F7($\theta$), T4-Pz($\theta$), F4-P4($\beta$), Fp2-T4($\theta$), Fz-C4($\delta$), F8-Cz($\delta$), T4-Pz($\gamma$), Fp1-F8($\delta$)  &  0.934  & Fp1-F7($\theta$), P4-O1($\theta$), T4-Pz($\theta$), F8-T3($\gamma$), Fp1-F4($\beta$), F7-P4($\gamma$), F4-C4($\delta$), Cz-T6($\beta$), C3-T4($\beta$)  &  0.875 \\   \hline
 10  & P4-O1($\theta$), Fp1-F7($\theta$), T4-Pz($\theta$), F4-P4($\beta$), Fp2-T4($\theta$), Fz-C4($\delta$), F8-Cz($\delta$), T4-Pz($\gamma$), Fp1-F8($\delta$), T3-P3($\gamma$)  &  0.907  & Fp1-F7($\theta$), P4-O1($\theta$), T4-Pz($\theta$), F8-T3($\gamma$), Fp1-F4($\beta$), F7-P4($\gamma$), F4-C4($\delta$), Cz-T6($\beta$), F8-T4($\beta$), C3-T4($\beta$)  &  0.836 \\   \hline
 11  & P4-O1($\theta$), Fp1-F7($\theta$), T4-Pz($\theta$), F4-P4($\beta$), Fp2-T4($\theta$), Fz-C4($\delta$), F8-Cz($\delta$), T4-Pz($\gamma$), Fp1-F8($\delta$), T3-P3($\gamma$), Fp1-F4($\beta$)  &  0.907  & Fp1-F7($\theta$), P4-O1($\theta$), T4-Pz($\theta$), F8-T3($\gamma$), Fp1-F4($\beta$), F7-P4($\gamma$), F4-C4($\delta$), Cz-T6($\beta$), F8-T4($\beta$), C3-T4($\beta$), C3-P3($\alpha$)  &  0.873 \\   \hline
 12  & P4-O1($\theta$), Fp1-F7($\theta$), T4-Pz($\theta$), F4-P4($\beta$), Fp2-T4($\theta$), Fz-C4($\delta$), F8-Cz($\delta$), T4-Pz($\gamma$), Fp1-F8($\delta$), T3-P3($\gamma$), Fp1-F4($\beta$), C3-O1($\delta$)  &  0.879  & Fp1-F7($\theta$), P4-O1($\theta$), T4-Pz($\gamma$), T4-Pz($\theta$), F8-T3($\gamma$), Fp1-F4($\beta$), F7-P4($\gamma$), F4-C4($\delta$), Cz-T6($\beta$), F8-T4($\beta$), C3-T4($\beta$), C3-P3($\alpha$)  &  0.873 \\   \hline
 13  & P4-O1($\theta$), Fp1-F7($\theta$), T4-Pz($\theta$), F4-P4($\beta$), Fp2-T4($\theta$), Fz-C4($\delta$), F8-Cz($\delta$), T4-Pz($\gamma$), Fp1-F8($\delta$), T3-P3($\gamma$), Fp1-F4($\beta$), C3-O1($\delta$), Fp1-F7($\alpha$)  &  0.893  & Fp1-F7($\theta$), P4-O1($\theta$), T4-Pz($\gamma$), T4-Pz($\theta$), F4-C3($\alpha$), F8-T3($\gamma$), Fp1-F4($\beta$), F7-P4($\gamma$), F4-C4($\delta$), Cz-T6($\beta$), F8-T4($\beta$), C3-T4($\beta$), C3-P3($\alpha$)  &  0.888 \\   \hline
 14  & P4-O1($\theta$), Fp1-F7($\theta$), T4-Pz($\theta$), F4-P4($\beta$), Fp2-T4($\theta$), Fz-C4($\delta$), F8-Cz($\delta$), T4-Pz($\gamma$), Fp1-F8($\delta$), T3-P3($\gamma$), Fp1-F4($\beta$), C3-O1($\delta$), Fp1-F7($\alpha$), F7-P4($\gamma$)  &  0.879  & Fp1-F7($\theta$), P4-O1($\theta$), T4-Pz($\gamma$), T4-Pz($\theta$), F4-C4($\theta$), F4-C3($\alpha$), F8-T3($\gamma$), Fp1-F4($\beta$), F7-P4($\gamma$), F4-C4($\delta$), Cz-T6($\beta$), F8-T4($\beta$), C3-T4($\beta$), C3-P3($\alpha$)  &  0.875 \\   \hline
 15  & P4-O1($\theta$), Fp1-F7($\theta$), T4-Pz($\theta$), F4-P4($\beta$), Fp2-T4($\theta$), Fz-C4($\delta$), F8-Cz($\delta$), T4-Pz($\gamma$), Fp1-F8($\delta$), T3-P3($\gamma$), Fp1-F4($\beta$), C3-O1($\delta$), Fp1-F7($\alpha$), F7-P4($\gamma$), T5-O1($\theta$)  &  0.907  & Fp1-F7($\theta$), P4-O1($\theta$), T4-Pz($\gamma$), T4-Pz($\theta$), F4-C4($\theta$), Fz-P4($\gamma$), F4-C3($\alpha$), F8-T3($\gamma$), Fp1-F4($\beta$), F7-P4($\gamma$), F4-C4($\delta$), Cz-T6($\beta$), F8-T4($\beta$), C3-T4($\beta$), C3-P3($\alpha$)  &  0.852 \\   \hline
 16  & P4-O1($\theta$), Fp1-F7($\theta$), T4-Pz($\theta$), F4-P4($\beta$), F8-Cz($\delta$), Fp1-F4($\beta$), T5-O1($\theta$), Fz-P4($\gamma$), F8-T4($\beta$), F3-F4($\alpha$), C3-T4($\beta$), F4-F8($\alpha$), Fz-F4($\alpha$), F8-T5($\delta$), Fz-Cz($\delta$), Fp2-F7($\theta$)  &  0.948  & Fp1-F7($\theta$), P4-O1($\theta$), T4-Pz($\gamma$), T4-Pz($\theta$), F4-C4($\theta$), Fz-P4($\gamma$), F4-C3($\alpha$), F8-T3($\gamma$), Fp1-F4($\beta$), F7-P4($\gamma$), F4-C4($\delta$), Cz-T6($\beta$), F4-P4($\beta$), F8-T4($\beta$), C3-T4($\beta$), C3-P3($\alpha$)  &  0.777 \\   \hline
 17  & P4-O1($\theta$), Fp1-F7($\theta$), T4-Pz($\theta$), F4-P4($\beta$), F8-Cz($\delta$), Fp1-F4($\beta$), T5-O1($\theta$), F8-T3($\gamma$), Fz-P4($\gamma$), F8-T4($\beta$), F3-F4($\alpha$), C3-T4($\beta$), F4-F8($\alpha$), Fz-F4($\alpha$), F8-T5($\delta$), Fz-Cz($\delta$), Fp2-F7($\theta$)  &  0.948  & Fp1-F7($\theta$), P4-O1($\theta$), T4-Pz($\gamma$), T4-Pz($\theta$), F4-C4($\theta$), Fp2-T4($\theta$), Fz-P4($\gamma$), F4-C3($\alpha$), F8-T3($\gamma$), Fp1-F4($\beta$), F7-P4($\gamma$), F4-C4($\delta$), Cz-T6($\beta$), F4-P4($\beta$), F8-T4($\beta$), C3-T4($\beta$), F8-Cz($\delta$)  &  0.877 \\   \hline
 18  & P4-O1($\theta$), Fp1-F7($\theta$), T4-Pz($\theta$), F4-P4($\beta$), F8-Cz($\delta$), Fp1-F4($\beta$), T5-O1($\theta$), F8-T3($\gamma$), Fz-P4($\gamma$), F8-T4($\beta$), F3-F4($\alpha$), T3-C3($\gamma$), C3-T4($\beta$), F4-F8($\alpha$), Fz-F4($\alpha$), F8-T5($\delta$), Fz-Cz($\delta$), Fp2-F7($\theta$)  &  0.934  & Fp1-F7($\theta$), P4-O1($\theta$), T4-Pz($\gamma$), T4-Pz($\theta$), F4-C4($\theta$), Fp2-T4($\theta$), Fz-P4($\gamma$), F4-C3($\alpha$), F8-T3($\gamma$), T3-C3($\gamma$), Fp1-F4($\beta$), F7-P4($\gamma$), F4-C4($\delta$), Cz-T6($\beta$), F4-P4($\beta$), F8-T4($\beta$), C3-T4($\beta$), F8-Cz($\delta$)  &  0.850 \\   \hline
 19  & P4-O1($\theta$), Fp1-F7($\theta$), T4-Pz($\theta$), F4-P4($\beta$), F8-Cz($\delta$), Fp1-F4($\beta$), T5-O1($\theta$), T5-P3($\delta$), F8-T3($\gamma$), Fz-P4($\gamma$), F8-T4($\beta$), F3-F4($\alpha$), T3-C3($\gamma$), C3-T4($\beta$), F4-F8($\alpha$), Fz-F4($\alpha$), F8-T5($\delta$), Fz-Cz($\delta$), Fp2-F7($\theta$)  &  0.961  & Fp1-F7($\theta$), P4-O1($\theta$), T4-Pz($\gamma$), T4-Pz($\theta$), F4-C4($\theta$), Fp2-T4($\theta$), Fz-P4($\gamma$), F4-C3($\alpha$), F8-T3($\gamma$), T3-C3($\gamma$), Fz-F8($\delta$), Fp1-F4($\beta$), F7-P4($\gamma$), F4-C4($\delta$), Cz-T6($\beta$), F4-P4($\beta$), F8-T4($\beta$), C3-T4($\beta$), F8-Cz($\delta$)  &  0.864 \\   \hline
 20  & P4-O1($\theta$), Fp1-F7($\theta$), T4-Pz($\theta$), F4-P4($\beta$), F8-Cz($\delta$), Fp1-F4($\beta$), T5-O1($\theta$), T3-T6($\delta$), T5-P3($\delta$), F8-T3($\gamma$), Fz-P4($\gamma$), F8-T4($\beta$), F3-F4($\alpha$), T3-C3($\gamma$), C3-T4($\beta$), F4-F8($\alpha$), Fz-F4($\alpha$), F8-T5($\delta$), Fz-Cz($\delta$), Fp2-F7($\theta$)  &  0.975  & Fp1-F7($\theta$), P4-O1($\theta$), T4-Pz($\gamma$), T4-Pz($\theta$), F4-C4($\theta$), Fp2-T4($\theta$), Fz-P4($\gamma$), F4-C3($\alpha$), F8-T3($\gamma$), T3-C3($\gamma$), Fz-F8($\delta$), Fp1-F4($\beta$), F7-P4($\gamma$), F4-C4($\delta$), C3-O1($\delta$), Cz-T6($\beta$), F4-P4($\beta$), F8-T4($\beta$), C3-T4($\beta$), F8-Cz($\delta$)  &  0.836 \\  \hline 
\end{tabular}
    \caption{Evolution of feature selection for mSFFS and SFFS when the number of features $k$ increases from 1 to 20. }
    \label{tab:feature_evolving}
\end{table*}

\section{Results}
\label{sec:results}
\subsection{Selected features}
We produced feature selections of increasing dimensionality (from 1 to 50) and obtained their corresponding cross-validation scores (Fig.~\ref{fig:msffs_curve}) and test accuracies (Fig.~\ref{fig:fs_compare}). According to the cross-validation curve, using 20 features produces the best mean validation score of 0.960 with a relatively small standard deviation of 0.018. This happens to be where the mean testing score peaks (0.975). We, therefore, use these 20 features as the final selected connectivity features. They are: P4-O1 theta, Fp1-F7 theta, T4-Pz theta, F4-P4 beta, F8-Cz delta, Fp1-F4 beta, T5-O1 theta, T3-T6 delta, T5-P3 delta, F8-T3 gamma, Fz-P4 gamma, F8-T4 beta, F3-F4 alpha, T3-C3 gamma, C3-T4 beta, F4-F8 alpha, Fz-F4 alpha, F8-T5 delta, Fz-Cz delta, and Fp2-F7 theta. 

\paragraph{Evolution of feature selections}
To gain some insight on the floating selection processes in both SFFS and mSFFS and how the selection outcomes evolve, the selected feature sets are displayed in Table~\ref{tab:feature_evolving} when the number of features increases from 1 to 20. Starting from the same first two chosen features, the feature selections of both schemes start to diverge from $k=4$, yet maintaining some similarity to each other. 

\begin{figure}[!h]
  \centerline{\includegraphics[width=.9\columnwidth]{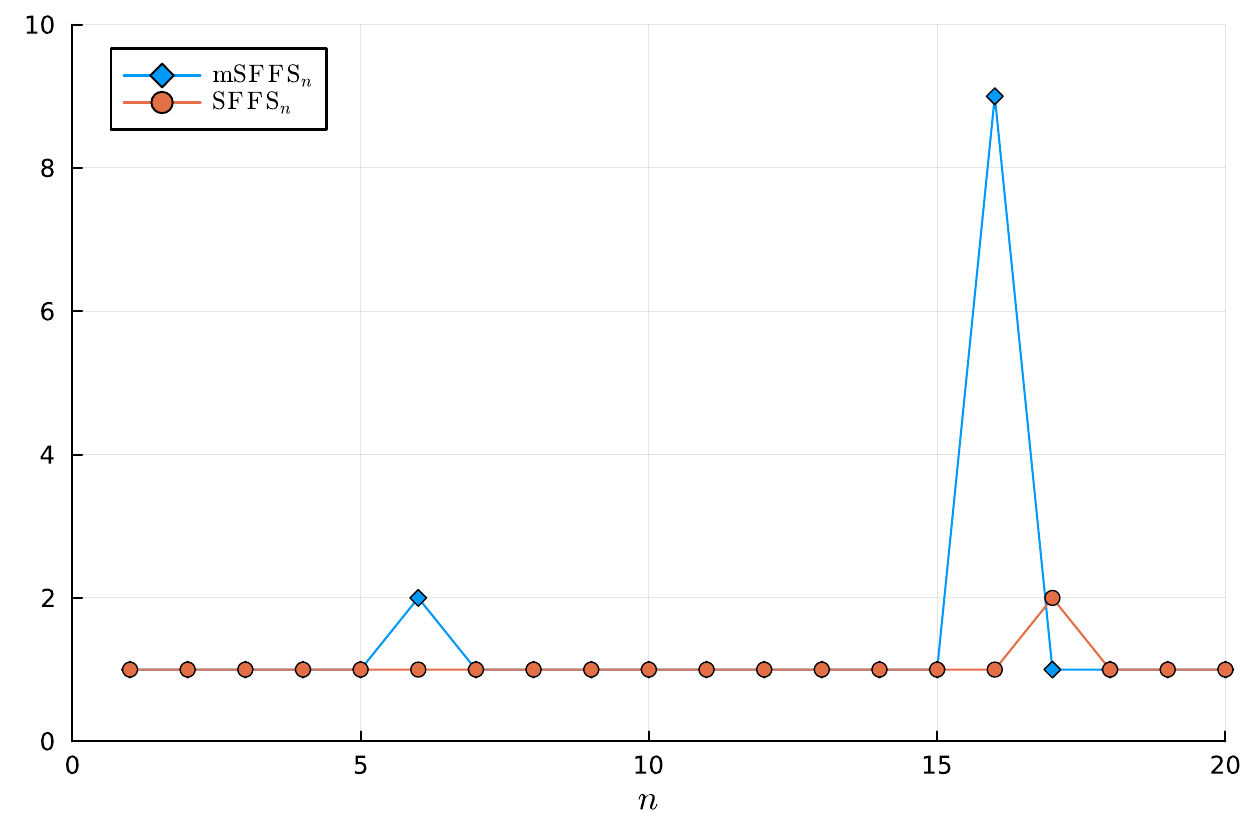}}  
  \caption{Incremental differences of feature selections by SFFS and mSFFS.}
  \label{fig:feat_inc}
\end{figure}

Fig.~\ref{fig:feat_inc} sheds some light on how the search paths of SFFS and mSFFS differ over time. As can be seen, for $n=1,\cdots,20$, SFFS mainly follows incremental addition of the next best feature, with $n=17$ being the only exception where this linear growth pattern is disrupted, and two new features get selected for that round. For mSFFS, there is a similar, small disruption at $n=6$, and also a major disruption at $n=16$, where 9 new features make into the selection. This indicates that mSFFS seems more capable of escaping from local-minimum traps and branch out new search paths than SFFS. 

Correlating this behaviour with the validation and testing performance trends as shown in Figs.~\ref{fig:msffs_curve} and ~\ref{fig:fs_compare}, we can see that $n=16$ is where mSFFS starts to gain a visible performance margin against SFFS, before reaching the performance peak at $n=20$. 
\begin{figure}[!h]
  \centerline{\includegraphics[width=.9\columnwidth]{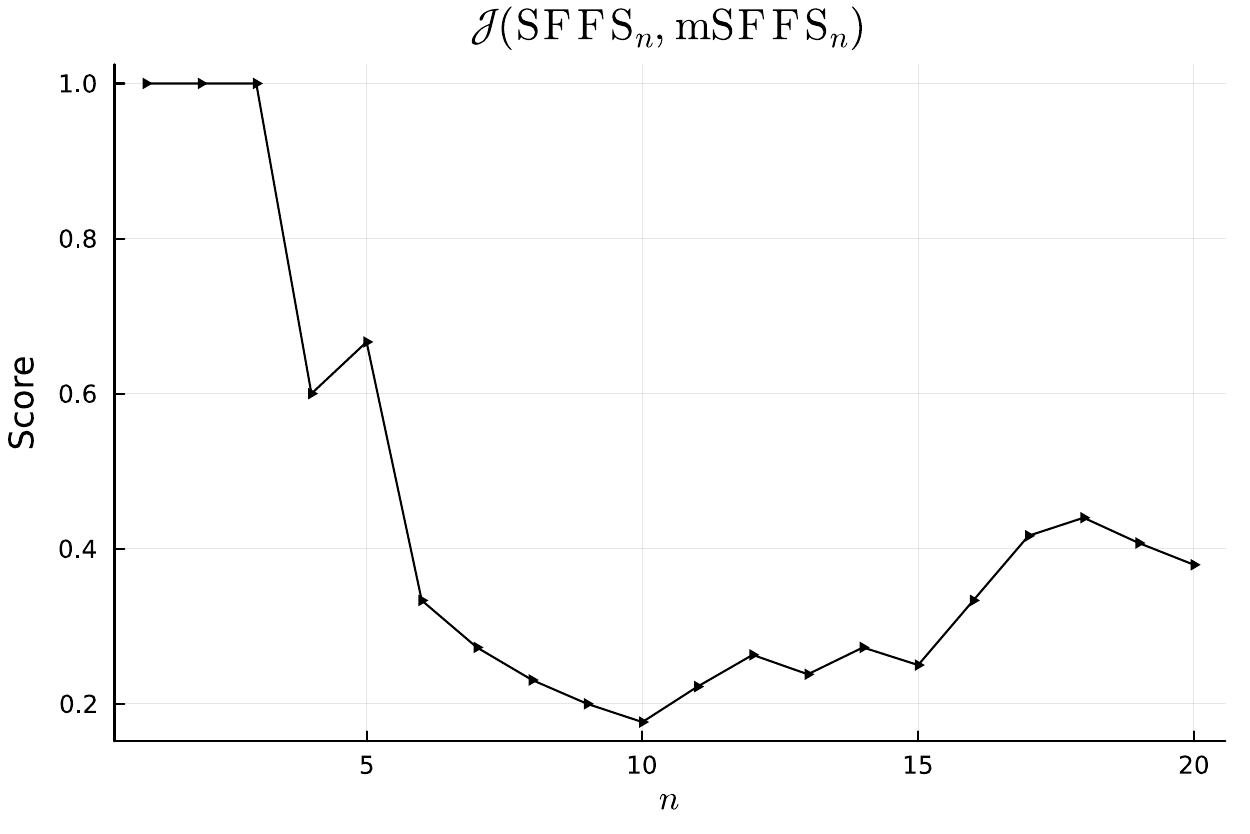}}  
  \caption{Jaccard index of the two evolving feature selections over time.}
  \label{fig:jac_n}
\end{figure}
\paragraph{Jaccard indeces} 
We further employ the Jaccard index between feature selection outcomes to quantify the agreement between two feature schemes. As shown in Fig.~\ref{fig:jac_n}, when $n$ increases from 1 to 20, the two feature schemes deviate from initial full overlap ($\mathcal{J}(n)=1, n=1,2,3$), drop to the lowest score at $n=10$, before developing some moderate overlap again.  

Upon reaching $k=20$, the mSFFS selection gives a Jaccard index of 0.379 to that of SFFS. It overlaps less with the top-20 (with the highest SHAP values) feature set, with a Jaccard index of 0.333, while the same index between SFFS and top-20 selections is 0.538. Hence mSFFS's search route further deviates from the top-20 set, allowing alternative features to be explored for inclusion that have less redundancy to existing features but larger contributions to classification. 

\begin{figure}[!t]
\centering
\begin{subfigure}{0.7\columnwidth}
\includegraphics[width=1\linewidth]{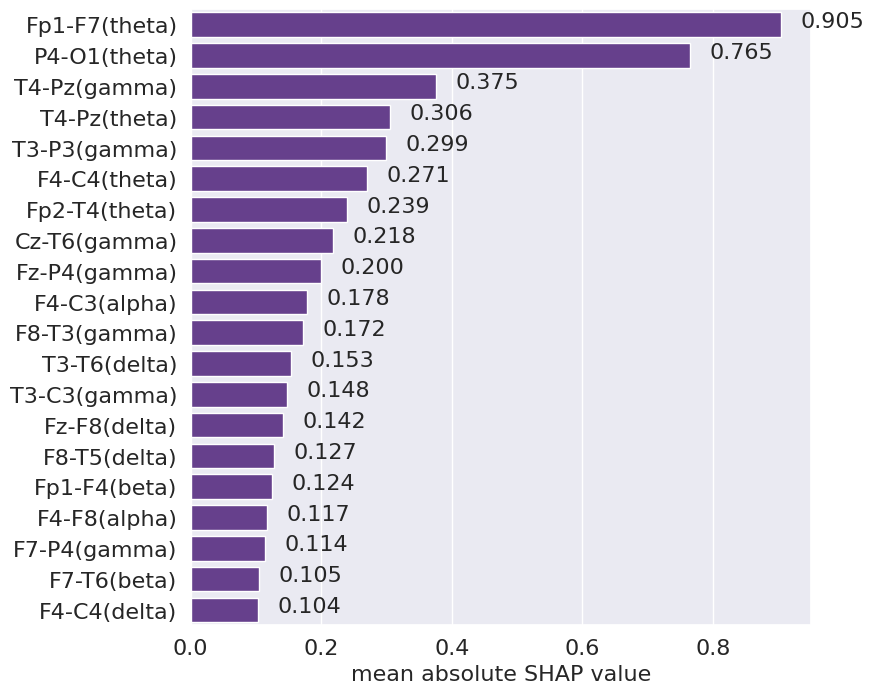}
\caption{SHAP}
\label{fig:SHAP_topf}
\end{subfigure}

\begin{subfigure}{0.7\columnwidth}
\includegraphics[width=1\linewidth]{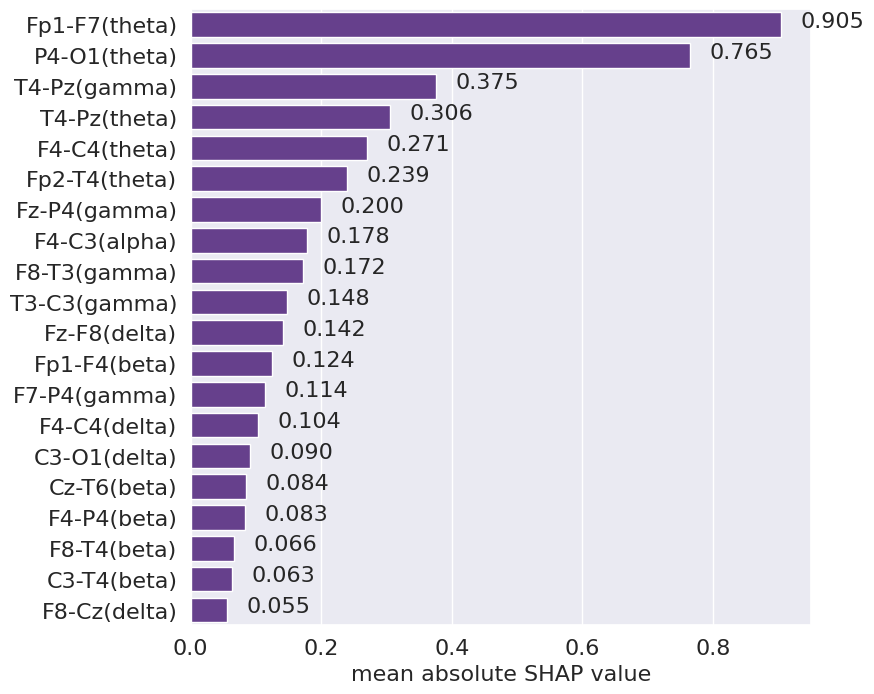}
\caption{SFFS}
\label{fig:SHAP_SFFS}
\end{subfigure}

\begin{subfigure}{0.7\columnwidth}
\includegraphics[width=1\linewidth]{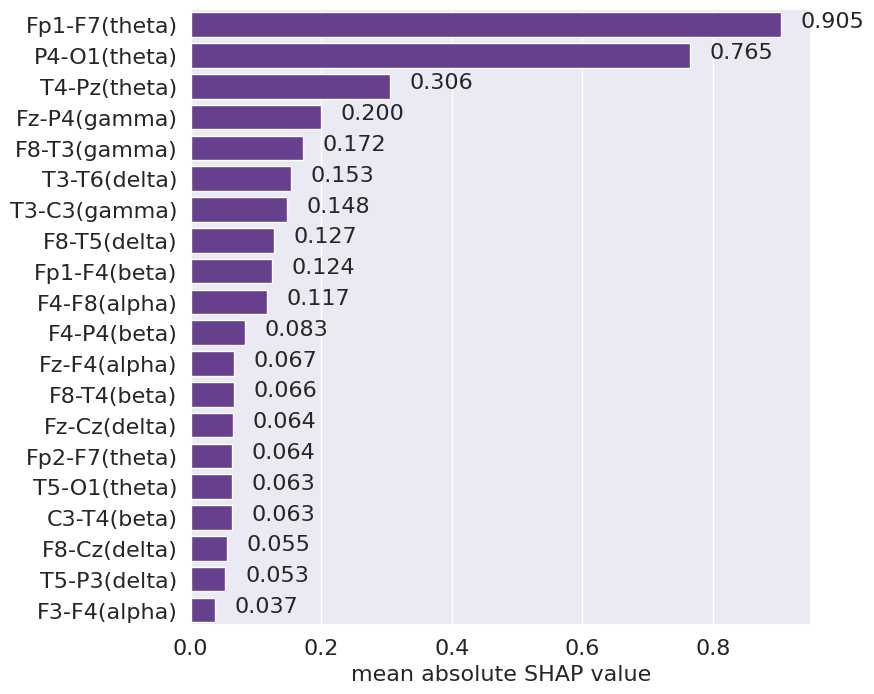}
\caption{mSFFS}
\label{fig:SHAP_mSFFS}
\end{subfigure}

\caption{Mean absolute SHAP values of the features selected by three feature selection schemes: (a) top-20 SHAP; (b) SFFS, 20 features; (c) SFFS, 20 features.}
\label{fig:top-20sels}
\end{figure}

\paragraph{SHAP value comparison}
The sorted, mean absolute SHAP values of the features selected by three feature selection schemes are shown in Fig.~\ref{fig:top-20sels}. The difference in the search outcome in mSFFS is also reflected here. Clearly, mSFFS does not simply favor features with large SHAP values, which may help it overcome the simple greedy nature of SFFS, allowing \textit{small} features with lesser SHAP values to be chosen and eventually contribute to better performance.

\subsection{Classification performance}
We followed the nested cross-validation process described in Section ~\ref{sec:mt}.
The average cross-validation score on the training set is 0.960 ($\pm$0.018), and the average test score is 0.975 ($\pm$0.050), using the mSFFS scheme.


We compared the classification performance of the mSFFS scheme with SFFS, top-20 SHAP selections, and selections given by the swarm-based schemes, Boruta~\cite{kursa2010boruta}, and Minimum Redundancy and Maximum Relevance (mRMR)~\cite{peng2005feature}.  
None of the swarm-based feature selection schemes, including BPSO~\cite{tran2014overview}, BHH~\cite{thaher2020binary}, BGW~\cite{emary2016binary} and IDA~\cite{hammouri2020improved}, effectively reduced the feature dimension. SHAP and SFFS can designate a specific feature number, but their performance is worse than mSFFS.
Table~\ref{tab:perf} gives the number of features and test accuracy from each feature selection algorithm. It can be seen that swarm-based selection methods failed to find a compact feature selection, all resulting in dimensionalities larger than 400. The corresponding feature selections also performed poorly in classification. 
Boruta has reduced the number of features to nine, but the testing performance is inadequate. Other methods, including mRMR, SHAP and SFFS, can specify the number of features needed, but failed to produce optimal results.

\begin{table}[!t]
\scriptsize
    \centering
    \begin{tabular}{c|c|c|c|c|c|} 
    \hline
         \textbf{Algorithm} & \textbf{No.features} & \textbf{Accuracy} & \textbf{F1} & \textbf{\textcolor{black}{Precision}} & \textbf{\textcolor{black}{Recall}} \\ \hline
         BPSO & 417 & 0.607 & \textcolor{black}{0.501} & \textcolor{black}{0.617} & \textcolor{black}{0.508}\\
         BHH & 556 & 0.539 & \textcolor{black}{0.478} & \textcolor{black}{0.511} & \textcolor{black}{0.508} \\
         BGW & 615 & 0.552 & \textcolor{black}{0.468} & \textcolor{black}{0.532} & \textcolor{black}{0.450} \\
         IDA & 404 & 0.602 & \textcolor{black}{0.532} & \textcolor{black}{0.600} & \textcolor{black}{0.558} \\ 
         \textcolor{black}{Boruta} & \textcolor{black}{9} & \textcolor{black}{0.784} & \textcolor{black}{0.807} & \textcolor{black}{0.757} & \textcolor{black}{0.875} \\
         \textcolor{black}{mRMR} & \textcolor{black}{20} & \textcolor{black}{0.864} & \textcolor{black}{0.869} & \textcolor{black}{0.847} & \textcolor{black}{0.917} \\
         SHAP & 20 & 0.795 & \textcolor{black}{0.799} & \textcolor{black}{0.838} & \textcolor{black}{0.808} \\
         SFFS & 20 & 0.836 & \textcolor{black}{0.858} & \textcolor{black}{0.818} & \textcolor{black}{0.925} \\
         mSFFS & 20 & 0.975 & \textcolor{black}{0.892} & \textcolor{black}{0.922} & \textcolor{black}{0.875} \\ \hline
    \end{tabular}
    \caption{Feature selection schemes and their testing performance in  comparison.}
    \label{tab:perf}
\end{table}
\begin{table}[!t]
\scriptsize
    \centering
    \begin{tabular}{c||c|c||c|c}
\hline
\multirow{2}{*}{\textbf{$k$}} & \multicolumn{2}{c||}{\textbf{mSFFS vs SFFS}} & \multicolumn{2}{c}{\textbf{mSFFS vs SHAP}} \\ \cline{2-5} 
 	&	t-statistics	&	$p$-value	&	t-statistics	&	$p$-value	\\ \hline
1	&	0.00	&	1.00	&	1.36	&	0.19	\\
2	&	0.00	&	1.00	&	0.00	&	1.00	\\
3	&	0.00	&	1.00	&	1.09	&	0.29	\\
4	&	-0.14	&	0.89	&	1.09	&	0.29	\\
5	&	0.67	&	0.51	&	0.77	&	0.45	\\
6	&	0.68	&	0.51	&	2.12	&	\textit{0.05}$^*$	\\
7	&	1.75	&	0.10	&	2.55	&	\textit{0.02}$^*$	\\
8	&	0.70	&	0.49	&	1.22	&	0.24	\\
9	&	1.03	&	0.32	&	2.07	&	\textit{0.05}$^*$	\\
10	&	1.24	&	0.23	&	1.20	&	0.24	\\
11	&	1.60	&	0.56	&	1.39	&	0.18	\\
12	&	0.09	&	0.93	&	0.23	&	0.82	\\
13	&	0.09	&	0.93	&	0.56	&	0.58	\\
14	&	0.07	&	0.95	&	0.00	&	1.00	\\
15	&	1.06	&	0.30	&	0.81	&	0.43	\\
\textbf{16}	&	2.69	&	\textit{0.02}$^*$	&	3.64	&	\textit{0.00}$^*$	\\
17	&	1.81	&	0.09	&	3.95	&	\textit{0.00}$^*$	\\
\textbf{18}	&	2.15	&	\textit{0.05}$^*$	&	2.88	&	\textit{0.01}$^*$	\\
\textbf{19}	&	2.63	&	\textit{0.03}$^*$	&	3.77	&	\textit{0.00}$^*$	\\
\textbf{20}	&	4.32	&	\textit{0.00}$^*$	&	3.90	&	\textit{0.00}$^*$	\\
21	&	1.31	&	0.21	&	2.83	&	\textit{0.01}$^*$	\\
\textbf{22}	&	2.24	&	\textit{0.04}$^*$	&	3.70	&	\textit{0.00}$^*$	\\
23	&	1.81	&	0.09	&	3.00	&	\textit{0.01}$^*$	\\
\textbf{24}	&	2.85	&	\textit{0.01}$^*$	&	2.66	&	\textit{0.02}$^*$	\\
25	&	1.28	&	0.22	&	3.03	&	\textit{0.01}$^*$	\\
\textbf{26}	&	2.62	&	\textit{0.02}$^*$	&	3.33	&	\textit{0.00}$^*$	\\
\textbf{27}	&	2.24	&	\textit{0.04}$^*$	&	2.99	&	\textit{0.01}$^*$	\\
\textbf{28}	&	2.16	&	\textit{0.04}$^*$	&	2.42	&	\textit{0.03}$^*$	\\
\textbf{29}	&	2.05	&	\textit{0.05}$^*$	&	2.19	&	\textit{0.04}$^*$	\\
30	&	1.76	&	0.09	&	1.75	&	0.10	\\  \hline  
\end{tabular}
    \caption{T-test results on the testing scores obtained from mSFFS selections versus those by SFFS and top SHAP values over a range of selected feature numbers ($k$) up to 30. Further results are statistical insignificance, hence omitted. Dimensionalities on which mSFFS outperforms both SFFS and SHAP with statistical significance ($p\le 0.05$) are highlighted in bold. Asterisks after $p$-values in italics indicate significance. }
    \label{tab:ttest}
\end{table}

\begin{figure*}[!t]
\centerline{
\begin{subfigure}{0.35\textwidth}
\includegraphics[width=\textwidth]{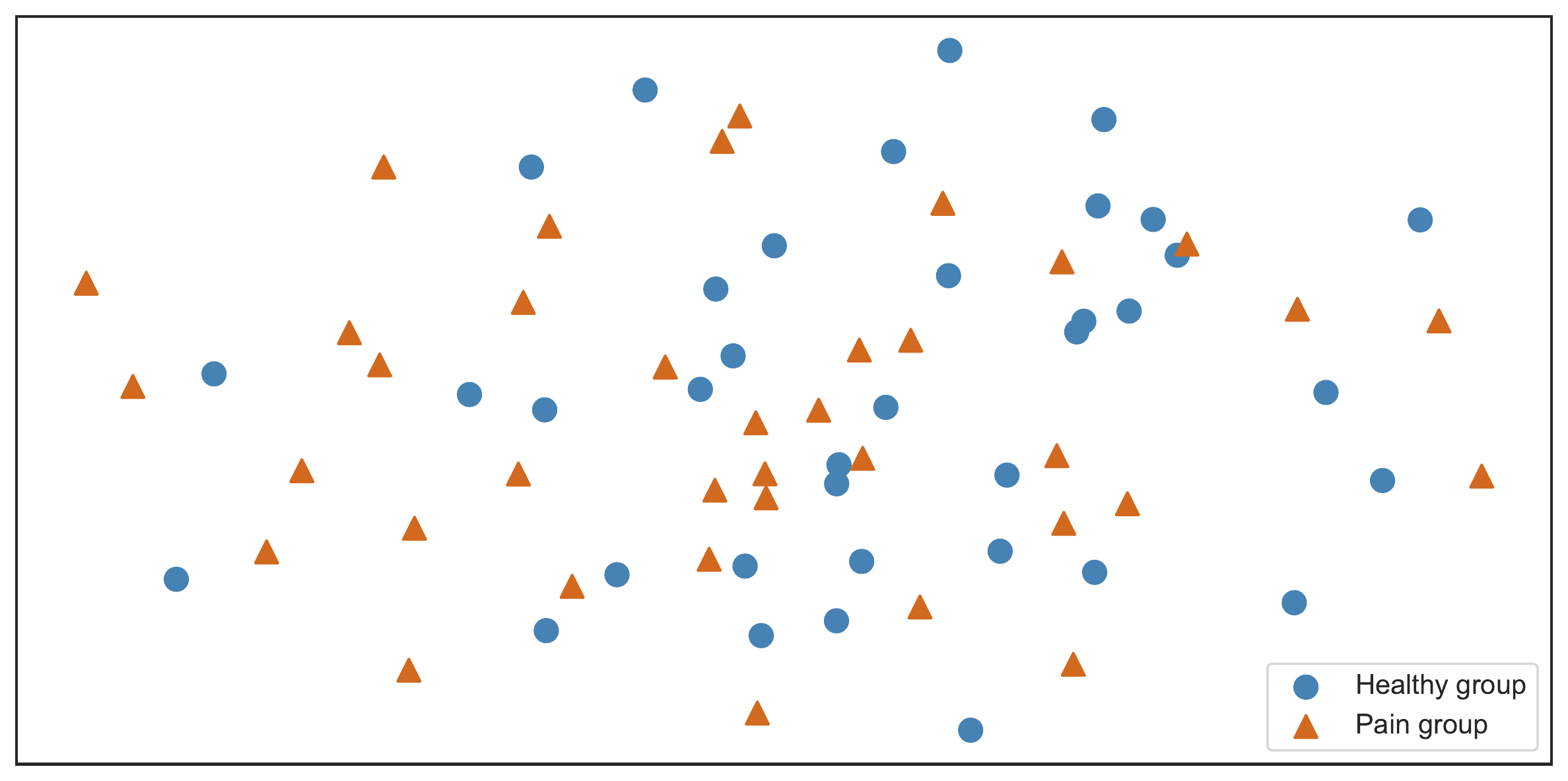}
\caption{}
\label{fig:tSNE_allf}
\end{subfigure}
\begin{subfigure}{0.35\textwidth}
\includegraphics[width=1\textwidth]{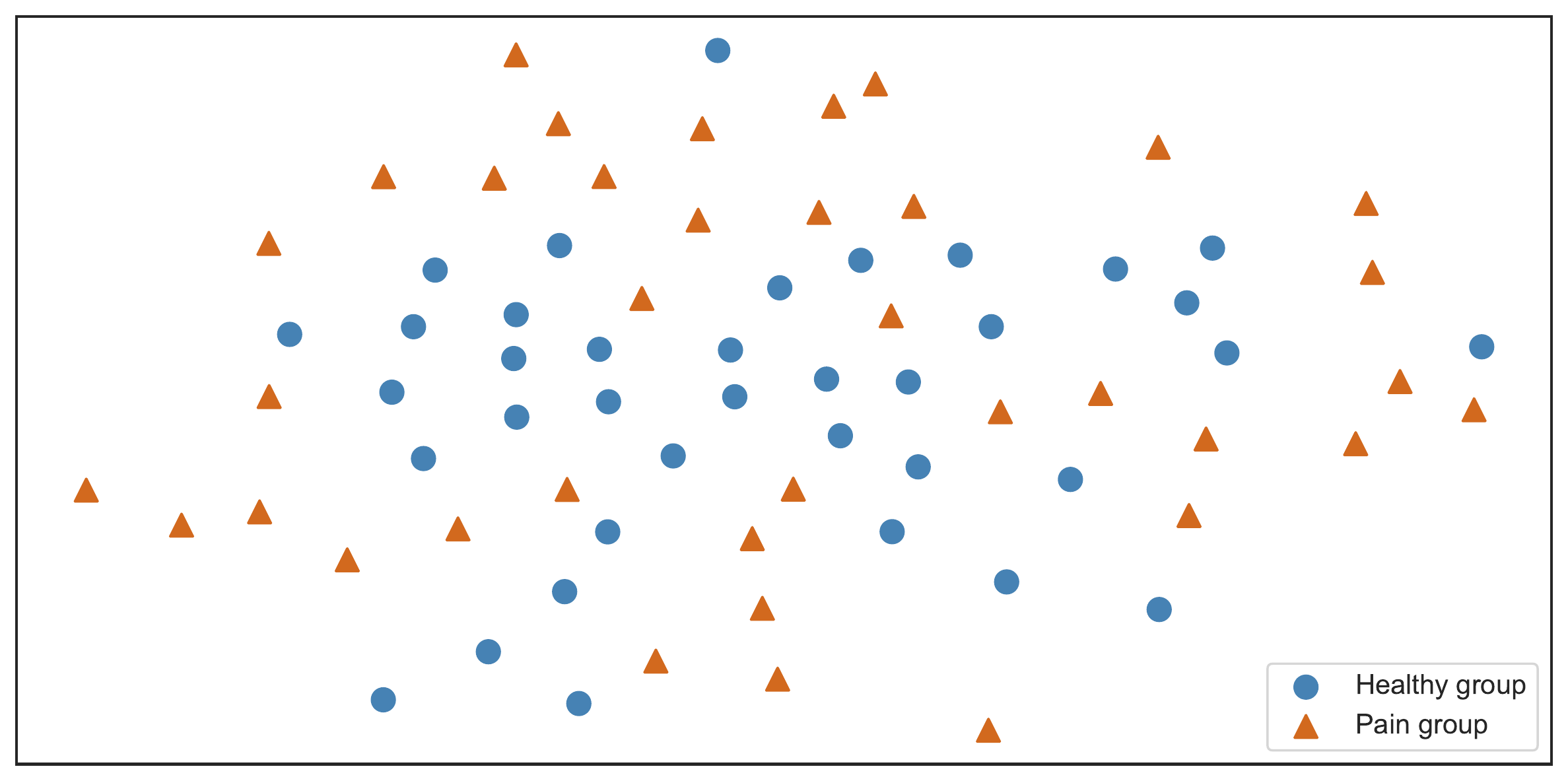}
\caption{}
\label{fig:tSNE_topf}
\end{subfigure}}

\centerline{
\begin{subfigure}{0.35\textwidth}
\includegraphics[width=1\linewidth]{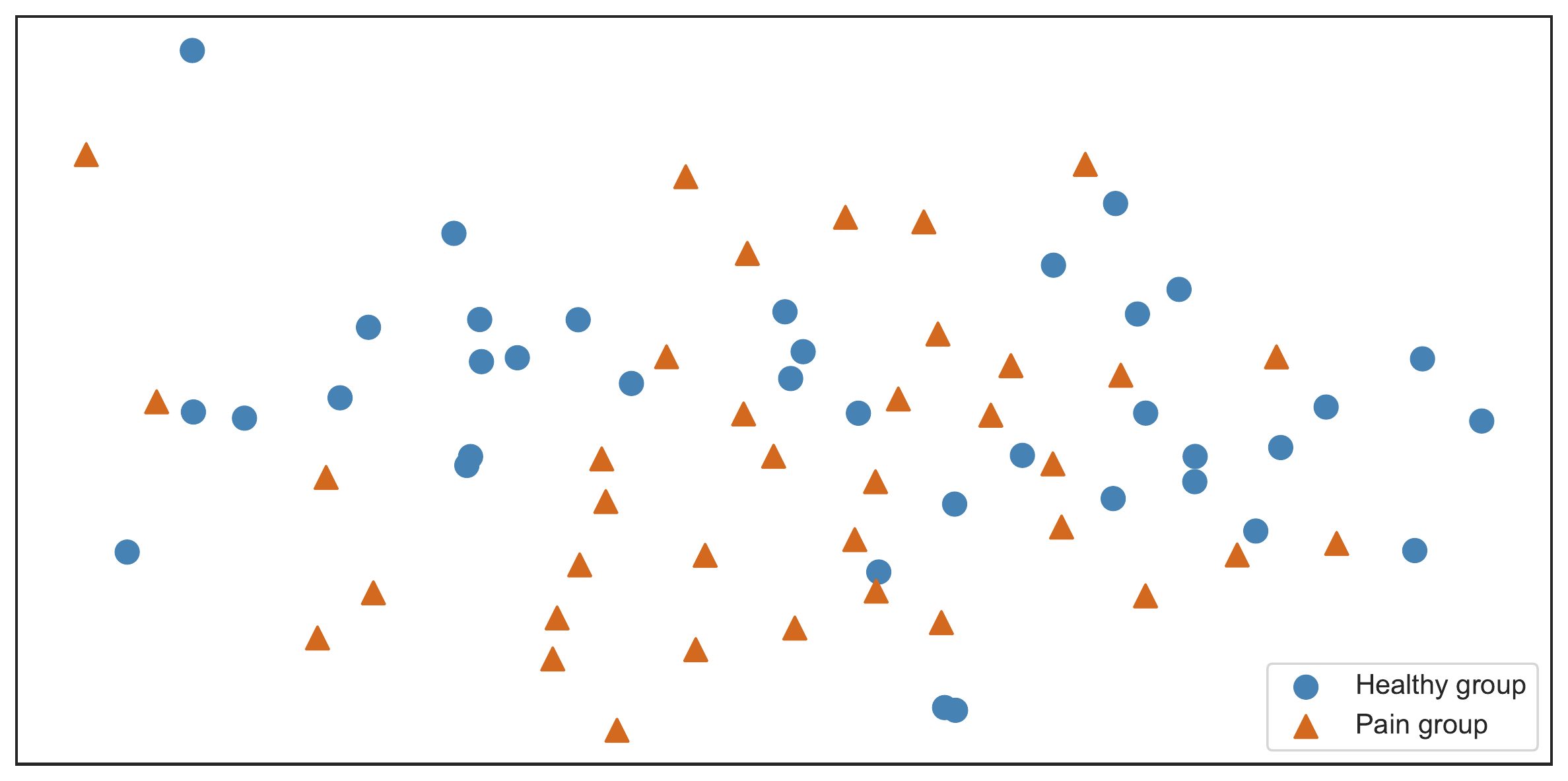}
\caption{}
\label{fig:tSNE_SFFSf}
\end{subfigure}
\begin{subfigure}{0.35\textwidth}
\includegraphics[width=1\linewidth]{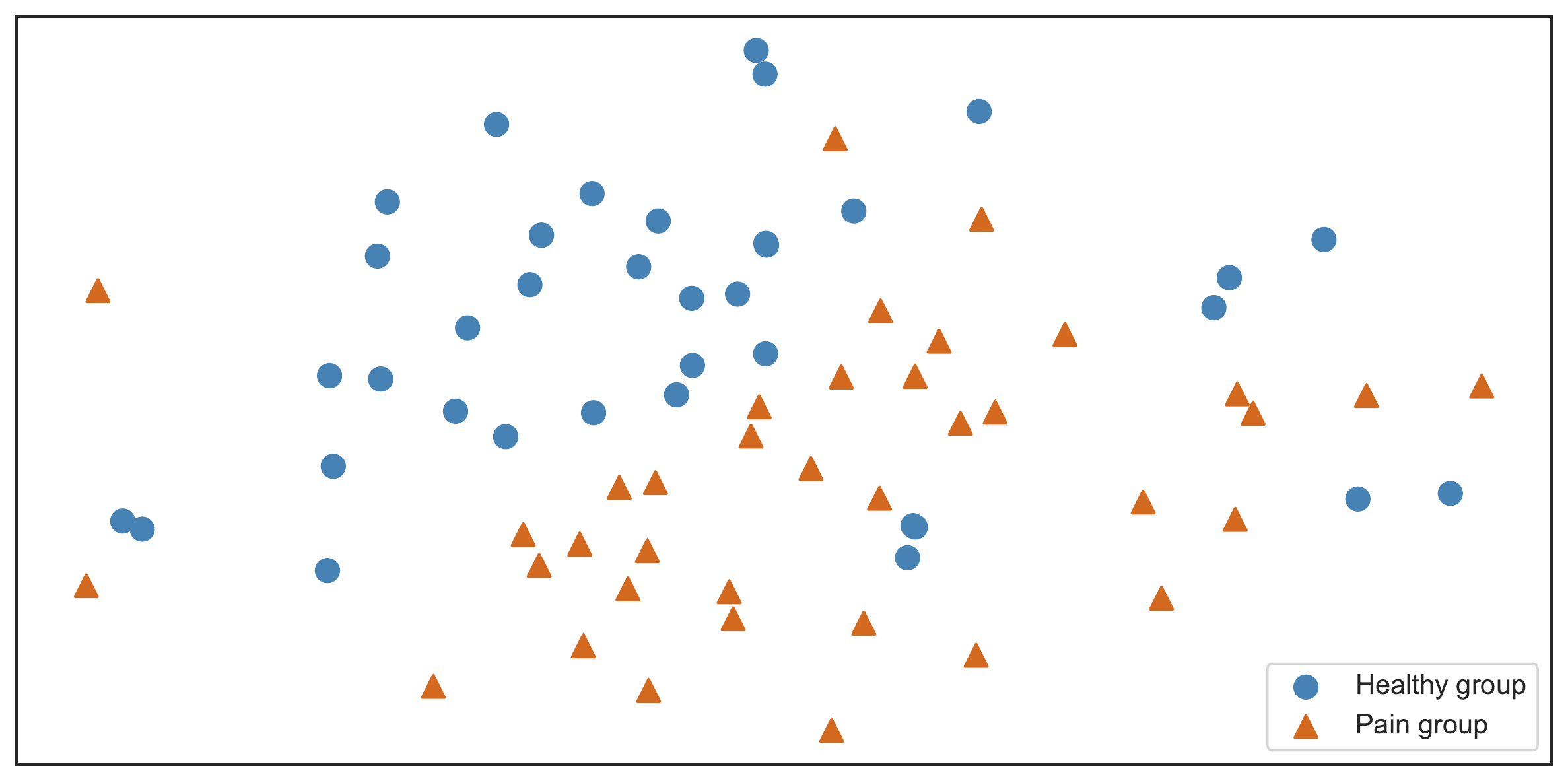}
\caption{}
\label{fig:tSNE_selectedf}
\end{subfigure}}
\caption{Outcome of t-SNE visualization of data distributions using: (a) all 855 features; (b) 20 top-ranked SHAP features; (c) 20 features selected by SFFS; (d) 20 features selected by mSFFS.}
\label{fig:tSNE}
\end{figure*}

\subsection{Detailed evaluations}

Further analyses demonstrated the differences between mSFFS and SFFS. In specific, t-test shows that the performance scores are significantly different; t-SNE illustrated a clear visual difference; and Bhattacharyya distance evaluation suggests that the classes are better separated using the mSFFS features. To gain the insight of how these two algorithms incorporate features with high importance, top-ranked SHAP is also included in the comparison. 

\paragraph{$t$-test}
We conducted two-sided $t$-tests to compare the test scores obtained from each scheme using 20 features. The results of the $t$-tests revealed that the comparison between mSFFS and SFFS had a $t$-statistics value of 4.316 and a $p$-value of 0.0004, indicating a statistically significant difference between the two schemes. Similarly, the comparison between mSFFS and SHAP selection yielded a statistics value of 3.896 with a $p$-value of 0.001, suggesting that there was also a significant difference between these two schemes. The $t$-test results over the range of selection dimensionality ($k=1\sim 30$) are detailed in Table~\ref{tab:ttest}. On 10 out of 30 occasions the performance gains of mSFFS are statistically significant against both SFFS and SHAP. 

\paragraph{t-SNE visualization}
The testing performance was aligned with the t-SNE visualization outcome. Figure~\ref{fig:tSNE} reveals the distribution of the two-class data samples before and after feature selection. Out of the three feature schemes (mSFFS, SFFS, and SHAP), the mSFFS selected features show the clearest separation between pain and healthy participants, which demonstrates the superiority of mSFFS in selecting features that contributes to differentiating the classes.

\paragraph{Bhattacharyya separabilities}
In addition to the visual assessment, we also measured the dissimilarity of the healthy and pain classes' probability distributions using the Bhattacharyya distance ($D_B$). This is accomplished using features selected through different schemes and embedded into a 2-dimensional space by t-SNE, followed by distance calculation. The result indicates that the data characterized by mSFFS features exhibit considerably greater segregation ($D_B=1.492$) between pain and healthy individuals when compared with SFFS ($D_B=1.256$) and SHAP-rank ($D_B=0.223$). 


To sum up, 
our results suggest that mSFFS is a more effective feature selection scheme than SFFS, SHAP-based selection, and state-of-the-art swarm-based algorithms.
In general, our proposed mSFFS scheme provides feature selections with better visualization and better separability, and can achieve higher accuracies with only a few features.


\subsection{Discussion}
Our basic approach in searching out optimal combinations of features is well-aligned with a network science approach to diagnosing pain in the brain~\cite{Deak2022-dv}. By exploring coherence or interaction among sensor nodes, our feature selection algorithm deviates from top-ranked individual features but searches for top-performing combinations, which is clearly shown in the search paths given in Table~\ref{tab:feature_evolving}, and the final feature compositions given in Fig.~\ref{fig:top-20sels}. Comparing SFFS and mSFFS, it seems that mSFFS searches deeper in the diverse feature space and deviates further from the plain top-20 selection. Although more weaker features are included in the mSFFS selection, the combination works the best and leads to the improvements as observed from visualization (Fig.~\ref{fig:tSNE}), and quantitative assessment on class separability and classification accuracy given in Section~\ref{sec:results}. 

The connectivity feature selection is identified in the sensor space using raw EEG signals instead of the source space, which  would often require manual data cleaning and extra processing for solving the inverse solution of retrieving the brain sources for generating the activity detected by the EEG sensors. The sensors that are involved however might be related to activity coming from somatosensory cortices (Pz and P4), i.e. the lateral pathway, and the anterior cingulate cortex (Fz and Cz), i.e. the medial pathway. The regions of the sensor signature we found overlap with some of the pain predictive brain regions found in other studies~\cite{kohoutova2022individual,Wager2013-nk,Vanneste2018}.
The activity in theta band (3 top features selected by mSFFS with sensors including frontal and right temporal sensors) is in agreement with the reported high theta band power in right frontal and temporal regions for subjects with high knee pain intensity~\cite{Simis2022}. Four beta band activities are selected into the 20-feature selection but with moderate importance, also partially overlapping (Fp1, F4, F8) with those reported beta activities~\cite{Simis2022}. 

It is interesting to see that sensors at regions related to emotional memory and cognitive processing (e.g., P4, T4 etc.) are included in the connectivity feature selection, even though these seem to be not reported in other studies. 

\section{Conclusion}
Despite being an important healthcare issue, chronic pain has received little data-centred treatment in the research literature. In this paper, we have demonstrated that 
a simple, cheap and clinically useful method of chronic knee pain prediction can perform on raw EEG data. This permits the development of a clinically applicable device that can detect and diagnose the presence of pain in an objective purely data-driven way with a test accuracy of 97.5\%.
This is highly relevant both from a research and clinical point of view. We consider the effectiveness and compactness of the selected functional connectivity a significant advantage, as this may potentially lead to cheap hardware implementations that can carry out automatic diagnosis of chronic knee pain, replacing the current, subjective self-reporting practice. The finding of a compact, reliable sensor-level connectivity signature for chronic knee pain could lead to potential neuromodulation applications for chronic knee pain relief and treatment. 

The performance gain we have achieved is due to the modified SFFS algorithm, whose better feature selection quality against SFFS and other state-of-the-art methods in comparison has been demonstrated from multiple perspectives: better cross validation performance, enhance external testing results, better t-SNE visualization outcome as well as improved quantitative class separability between the pain and healthy control groups. We have also analyzed the searching behaviour of the feature selection algorithms, which reveals that our algorithm tends to have more dynamics to escape from local minima and arrive at more competitive feature selections. 

We utilized EEG data acquired from two sources in this study. On the other hand, the so-called ``domain gaps'' were reported in the literature, namely, trained predictive models may have reduced generalization ability on EEGs acquired from diverse sources~\cite{Chai2017}. Another limitation of the current study is that we did not pursue a deep-learning solution owing to the small amount of data we had. 

For future work, we will seek to augment our datasets by including recordings from different sources. With data of larger amounts and more diversity available,  the subsequent investigation involves exploring deep domain adaption techniques~\cite{Hou:22,Wang:23} and developing classification models that perform steadily across acquisition sources.

\bibliographystyle{IEEEtran}
\bibliography{references}

\begin{thebibliography}{10}
\providecommand{\url}[1]{#1}
\csname url@samestyle\endcsname
\providecommand{\newblock}{\relax}
\providecommand{\bibinfo}[2]{#2}
\providecommand{\BIBentrySTDinterwordspacing}{\spaceskip=0pt\relax}
\providecommand{\BIBentryALTinterwordstretchfactor}{4}
\providecommand{\BIBentryALTinterwordspacing}{\spaceskip=\fontdimen2\font plus
\BIBentryALTinterwordstretchfactor\fontdimen3\font minus
  \fontdimen4\font\relax}
\providecommand{\BIBforeignlanguage}[2]{{%
\expandafter\ifx\csname l@#1\endcsname\relax
\typeout{** WARNING: IEEEtran.bst: No hyphenation pattern has been}%
\typeout{** loaded for the language `#1'. Using the pattern for}%
\typeout{** the default language instead.}%
\else
\language=\csname l@#1\endcsname
\fi
#2}}
\providecommand{\BIBdecl}{\relax}
\BIBdecl

\bibitem{Bonica}
J.~J. Bonica, ``The need of a taxonomy.'' \emph{Pain}, vol.~6, no.~3, pp.
  247--248, 1979.

\bibitem{Treede2019-gu}
R.-D. Treede, W.~Rief, A.~Barke, Q.~Aziz, M.~I. Bennett, R.~Benoliel, M.~Cohen,
  S.~Evers, N.~B. Finnerup, M.~B. First, M.~A. Giamberardino, S.~Kaasa,
  B.~Korwisi, E.~Kosek, P.~Lavand'homme, M.~Nicholas, S.~Perrot, J.~Scholz,
  S.~Schug, B.~H. Smith, P.~Svensson, J.~W.~S. Vlaeyen, and S.-J. Wang,
  ``Chronic pain as a symptom or a disease: the {IASP} classification of
  chronic pain for the international classification of diseases ({ICD}-11),''
  \emph{Pain}, vol. 160, no.~1, pp. 19--27, Jan. 2019.

\bibitem{Scholz2019-tw}
J.~Scholz, N.~B. Finnerup, N.~Attal, Q.~Aziz, R.~Baron, M.~I. Bennett,
  R.~Benoliel, M.~Cohen, G.~Cruccu, K.~D. Davis, S.~Evers, M.~First, M.~A.
  Giamberardino, P.~Hansson, S.~Kaasa, B.~Korwisi, E.~Kosek, P.~Lavand'homme,
  M.~Nicholas, T.~Nurmikko, S.~Perrot, S.~N. Raja, A.~S.~C. Rice, M.~C.
  Rowbotham, S.~Schug, D.~M. Simpson, B.~H. Smith, P.~Svensson, J.~W.~S.
  Vlaeyen, S.-J. Wang, A.~Barke, W.~Rief, R.-D. Treede, and {Classification
  Committee of the Neuropathic Pain Special Interest Group (NeuPSIG)}, ``The
  {IASP} classification of chronic pain for {ICD-11}: chronic neuropathic
  pain,'' \emph{Pain}, vol. 160, no.~1, pp. 53--59, Jan. 2019.

\bibitem{Kuner2016-wx}
R.~Kuner and H.~Flor, ``Structural plasticity and reorganisation in chronic
  pain,'' \emph{Nat. Rev. Neurosci.}, vol.~18, no.~1, pp. 20--30, Dec. 2016.

\bibitem{Mills2019-kr}
S.~E.~E. Mills, K.~P. Nicolson, and B.~H. Smith, ``Chronic pain: a review of
  its epidemiology and associated factors in population-based studies,''
  \emph{Br. J. Anaesth.}, vol. 123, no.~2, pp. e273--e283, Aug. 2019.

\bibitem{Moriarty2011-id}
O.~Moriarty, B.~E. McGuire, and D.~P. Finn, ``The effect of pain on cognitive
  function: a review of clinical and preclinical research,'' \emph{Prog.
  Neurobiol.}, vol.~93, no.~3, pp. 385--404, Mar. 2011.

\bibitem{Breivik2006-ya}
H.~Breivik, B.~Collett, V.~Ventafridda, R.~Cohen, and D.~Gallacher, ``Survey of
  chronic pain in {Europe}: prevalence, impact on daily life, and treatment,''
  \emph{Eur. J. Pain}, vol.~10, no.~4, pp. 287--333, May 2006.

\bibitem{Yongjun2020-zr}
Y.~Zheng, T.~Zhang, X.~Yang, Z.~Feng, F.~Qiu, G.~Xin, J.~Liu, F.~Nie, X.~Jin,
  and Y.~Liu, ``A survey of chronic pain in {China},'' \emph{Libyan J. Med.},
  vol.~15, no.~1, p. 1730550, Dec. 2020.

\bibitem{Mutubuki2020-iu}
E.~N. Mutubuki, Y.~Beljon, E.~T. Maas, F.~J. P.~M. Huygen, R.~W. J.~G. Ostelo,
  M.~W. van Tulder, and J.~M. van Dongen, ``The longitudinal relationships
  between pain severity and disability versus health-related quality of life
  and costs among chronic low back pain patients,'' \emph{Quality of Life
  Research}, vol.~29, no.~1, pp. 275--287, Jan. 2020.

\bibitem{GBD_2017}
{GBD 2017 Disease and Injury Incidence and Prevalence Collaborators}, ``Global,
  regional, and national incidence, prevalence, and years lived with disability
  for 354 diseases and injuries for 195 countries and territories, 1990-2017: a
  systematic analysis for the global burden of disease study 2017,''
  \emph{Lancet}, vol. 392, no. 10159, pp. 1789--1858, Nov. 2018.

\bibitem{Gaskin2012-wg}
D.~J. Gaskin and P.~Richard, ``The economic costs of pain in the {United
  States},'' \emph{Journal of Pain}, vol.~13, no.~8, pp. 715--724, Aug. 2012.

\bibitem{Van_Tulder1995-fk}
M.~W. van Tulder, B.~W. Koes, and L.~M. Bouter, ``A cost-of-illness study of
  back pain in the netherlands,'' \emph{Pain}, vol.~62, no.~2, pp. 233--240,
  Aug. 1995.

\bibitem{Dahlhamer2018-vn}
J.~Dahlhamer, J.~Lucas, C.~Zelaya, R.~Nahin, S.~Mackey, L.~DeBar, R.~Kerns,
  M.~Von~Korff, L.~Porter, and C.~Helmick, ``Prevalence of chronic pain and
  high-impact chronic pain among adults - {United States}, 2016,'' \emph{MMWR
  Morb. Mortal. Wkly. Rep.}, vol.~67, no.~36, pp. 1001--1006, Sep. 2018.

\bibitem{Jackson2015-dm}
T.~Jackson, S.~Thomas, V.~Stabile, X.~Han, M.~Shotwell, and K.~McQueen,
  ``Prevalence of chronic pain in low-income and middle-income countries: a
  systematic review and meta-analysis,'' \emph{Lancet}, vol. 385 Suppl 2, p.
  S10, Apr. 2015.

\bibitem{Wager2013-nk}
T.~D. Wager, L.~Y. Atlas, M.~A. Lindquist, M.~Roy, C.-W. Woo, and E.~Kross,
  ``An {fMRI-based} neurologic signature of physical pain,'' \emph{N. Engl. J.
  Med.}, vol. 368, no.~15, pp. 1388--1397, Apr. 2013.

\bibitem{Stucky2001-jg}
C.~L. Stucky, M.~S. Gold, and X.~Zhang, ``Mechanisms of pain,'' \emph{Proc.
  Natl. Acad. Sci. U. S. A.}, vol.~98, no.~21, pp. 11\,845--11\,846, Oct. 2001.

\bibitem{Jensen2014-uy}
M.~P. Jensen, M.~A. Day, and J.~Mir{\'o}, ``Neuromodulatory treatments for
  chronic pain: efficacy and mechanisms,'' \emph{Nat. Rev. Neurol.}, vol.~10,
  no.~3, pp. 167--178, Mar. 2014.

\bibitem{kim2021sex}
J.~A. Kim, R.~L. Bosma, K.~S. Hemington, A.~Rogachov, N.~R. Osborne, J.~C.
  Cheng, B.~T. Dunkley, and K.~D. Davis, ``Sex-differences in network level
  brain dynamics associated with pain sensitivity and pain interference,''
  \emph{Human Brain Mapping}, vol.~42, no.~3, pp. 598--614, 2021.

\bibitem{kim2020cross}
J.~A. Kim, R.~L. Bosma, K.~S. Hemington, A.~Rogachov, N.~R. Osborne, J.~C.
  Cheng, J.~Oh, B.~T. Dunkley, and K.~D. Davis, ``Cross-network coupling of
  neural oscillations in the dynamic pain connectome reflects chronic
  neuropathic pain in multiple sclerosis,'' \emph{NeuroImage: Clinical},
  vol.~26, p. 102230, 2020.

\bibitem{kim2021neural}
J.~A. Kim and K.~D. Davis, ``Neural oscillations: understanding a neural code
  of pain,'' \emph{The Neuroscientist}, vol.~27, no.~5, pp. 544--570, 2021.

\bibitem{De_Ridder2017-gv}
D.~De~Ridder, S.~Perera, and S.~Vanneste, ``State of the art: Novel
  applications for cortical stimulation,'' \emph{Neuromodulation}, vol.~20,
  no.~3, pp. 206--214, Apr. 2017.

\bibitem{Vanneste2018}
S.~Vanneste, J.-J. Song, and D.~De~Ridder, ``Thalamocortical dysrhythmia
  detected by machine learning,'' \emph{Nature Communications}, vol.~9, no.~1,
  pp. 1--13, 2018.

\bibitem{nezam2018novel}
T.~Nezam, R.~Boostani, V.~Abootalebi, and K.~Rastegar, ``A novel classification
  strategy to distinguish five levels of pain using the {EEG} signal
  features,'' \emph{IEEE Transactions on Affective Computing}, vol.~12, no.~1,
  pp. 131--140, 2018.

\bibitem{vuckovic2018prediction}
A.~Vuckovic, V.~J.~F. Gallardo, M.~Jarjees, M.~Fraser, and M.~Purcell,
  ``Prediction of central neuropathic pain in spinal cord injury based on {EEG}
  classifier,'' \emph{Clinical Neurophysiology}, vol. 129, no.~8, pp.
  1605--1617, 2018.

\bibitem{vijayakumar2017quantifying}
V.~Vijayakumar, M.~Case, S.~Shirinpour, and B.~He, ``Quantifying and
  characterizing tonic thermal pain across subjects from {EEG} data using
  random forest models,'' \emph{IEEE Transactions on Biomedical Engineering},
  vol.~64, no.~12, pp. 2988--2996, 2017.

\bibitem{hsiao2021machine}
F.-J. Hsiao, W.-T. Chen, L.-L.~H. Pan, H.-Y. Liu, Y.-F. Wang, S.-P. Chen, K.-L.
  Lai, and S.-J. Wang, ``Machine learning--based prediction of heat pain
  sensitivity by using resting-state eeg,'' \emph{Frontiers in
  Bioscience-Landmark}, vol.~26, no.~12, pp. 1537--1547, 2021.

\bibitem{okolo2018use}
C.~Okolo and A.~Omurtag, ``Use of dry electroencephalogram and support vector
  for objective pain assessment,'' \emph{Biomedical Instrumentation \&
  Technology}, vol.~52, no.~5, pp. 372--378, 2018.

\bibitem{jas2017autoreject}
M.~Jas, D.~A. Engemann, Y.~Bekhti, F.~Raimondo, and A.~Gramfort, ``Autoreject:
  Automated artifact rejection for meg and eeg data,'' \emph{NeuroImage}, vol.
  159, pp. 417--429, 2017.

\bibitem{perrin1987scalp}
F.~Perrin, O.~Bertrand, and J.~Pernier, ``Scalp current density mapping: value
  and estimation from potential data,'' \emph{IEEE Transactions on Biomedical
  Engineering}, no.~4, pp. 283--288, 1987.

\bibitem{perrin1989spherical}
F.~Perrin, J.~Pernier, O.~Bertrand, and J.~F. Echallier, ``Spherical splines
  for scalp potential and current density mapping,''
  \emph{Electroencephalography and Clinical Neurophysiology}, vol.~72, no.~2,
  pp. 184--187, 1989.

\bibitem{cohen2014analyzing}
M.~X. Cohen, \emph{Analyzing neural time series data: theory and
  practice}.\hskip 1em plus 0.5em minus 0.4em\relax MIT Press, 2014.

\bibitem{kayser2015benefits}
J.~Kayser and C.~E. Tenke, ``On the benefits of using surface laplacian
  (current source density) methodology in electrophysiology,''
  \emph{International journal of psychophysiology: official journal of the
  International Organization of Psychophysiology}, vol.~97, no.~3, p. 171,
  2015.

\bibitem{bruna2018phase}
R.~Bru{\~n}a, F.~Maest{\'u}, and E.~Pereda, ``Phase locking value revisited:
  teaching new tricks to an old dog,'' \emph{Journal of neural engineering},
  vol.~15, no.~5, p. 056011, 2018.

\bibitem{GramfortEtAl2013a}
A.~Gramfort, M.~Luessi, E.~Larson, D.~A. Engemann, D.~Strohmeier, C.~Brodbeck,
  R.~Goj, M.~Jas, T.~Brooks, L.~Parkkonen, and M.~S. H{\"a}m{\"a}l{\"a}inen,
  ``{{MEG}} and {{EEG}} data analysis with {{MNE}}-{{Python}},''
  \emph{Frontiers in Neuroscience}, vol.~7, no. 267, pp. 1--13, 2013.

\bibitem{Pudil1994floating}
P.~Pudil, J.~Novovi{\v{c}}ov{\'a}, and J.~Kittler, ``Floating search methods in
  feature selection,'' \emph{Pattern Recognition Letters}, vol.~15, no.~11, pp.
  1119--1125, 1994.

\bibitem{Lundberg_nips17}
S.~M. Lundberg and S.-I. Lee, ``A unified approach to interpreting model
  predictions,'' in \emph{Advances in Neural Information Processing Systems
  30}, I.~Guyon, U.~V. Luxburg, S.~Bengio, H.~Wallach, R.~Fergus,
  S.~Vishwanathan, and R.~Garnett, Eds.\hskip 1em plus 0.5em minus 0.4em\relax
  Curran Associates, Inc., 2017, pp. 4765--4774.

\bibitem{Chen:2016}
T.~Chen and C.~Guestrin, ``{XGBoost}: A scalable tree boosting system,'' in
  \emph{Proceedings of the 22nd ACM SIGKDD International Conference on
  Knowledge Discovery and Data Mining}, ser. KDD '16.\hskip 1em plus 0.5em
  minus 0.4em\relax New York, NY, USA: Association for Computing Machinery,
  2016, p. 785–794.

\bibitem{James2021}
G.~James, D.~Witten, T.~Hastie, and R.~Tibshirani, \emph{An Introduction to
  Statistical Learning: with Applications in R}, 2nd~ed.\hskip 1em plus 0.5em
  minus 0.4em\relax Springer, 2021.

\bibitem{raschkas_2018_mlxtend}
S.~Raschka, ``{MLxtend}: Providing machine learning and data science utilities
  and extensions to python’s scientific computing stack,'' \emph{The Journal
  of Open Source Software}, vol.~3, no.~24, Apr. 2018.

\bibitem{Xue:13}
B.~Xue, M.~Zhang, and W.~N. Browne, ``Particle swarm optimization for feature
  selection in classification: A multi-objective approach,'' \emph{IEEE
  Transactions on Cybernetics}, vol.~43, no.~6, pp. 1656--1671, 2013.

\bibitem{thaher2020binary}
T.~Thaher, A.~A. Heidari, M.~Mafarja, J.~S. Dong, and S.~Mirjalili, ``Binary
  harris hawks optimizer for high-dimensional, low sample size feature
  selection,'' \emph{Evolutionary Machine Learning Techniques: Algorithms and
  Applications}, pp. 251--272, 2020.

\bibitem{emary2016binary}
E.~Emary, H.~M. Zawbaa, and A.~E. Hassanien, ``Binary grey wolf optimization
  approaches for feature selection,'' \emph{Neurocomputing}, vol. 172, pp.
  371--381, 2016.

\bibitem{hammouri2020improved}
A.~I. Hammouri, M.~Mafarja, M.~A. Al-Betar, M.~A. Awadallah, and I.~Abu-Doush,
  ``An improved dragonfly algorithm for feature selection,''
  \emph{Knowledge-Based Systems}, vol. 203, p. 106131, 2020.

\bibitem{vandermaaten08}
L.~van~der Maaten and G.~Hinton, ``Visualizing data using {t-SNE},''
  \emph{Journal of Machine Learning Research}, vol.~9, no.~86, pp. 2579--2605,
  2008.

\bibitem{scikit-learn}
F.~Pedregosa, G.~Varoquaux, A.~Gramfort, V.~Michel, B.~Thirion, O.~Grisel,
  M.~Blondel, P.~Prettenhofer, R.~Weiss, V.~Dubourg, J.~Vanderplas, A.~Passos,
  D.~Cournapeau, M.~Brucher, M.~Perrot, and E.~Duchesnay, ``Scikit-learn:
  Machine learning in {P}ython,'' \emph{Journal of Machine Learning Research},
  vol.~12, pp. 2825--2830, 2011.

\bibitem{Choi2003-xk}
E.~Choi and C.~Lee, ``\BIBforeignlanguage{en}{Feature extraction based on the
  {Bhattacharyya} distance},'' \emph{\BIBforeignlanguage{en}{Pattern
  Recognition}}, vol.~36, no.~8, pp. 1703--1709, Aug. 2003.

\bibitem{Pandy22}
M.~Pándy, A.~Agostinelli, J.~Uijlings, V.~Ferrari, and T.~Mensink,
  ``Transferability estimation using {Bhattacharyya} class separability,'' in
  \emph{2022 IEEE/CVF Conference on Computer Vision and Pattern Recognition
  (CVPR)}, 2022, pp. 9162--9172.

\bibitem{kursa2010boruta}
M.~B. Kursa, A.~Jankowski, and W.~R. Rudnicki, ``Boruta--a system for feature
  selection,'' \emph{Fundamenta Informaticae}, vol. 101, no.~4, pp. 271--285,
  2010.

\bibitem{peng2005feature}
H.~Peng, F.~Long, and C.~Ding, ``Feature selection based on mutual information
  criteria of max-dependency, max-relevance, and min-redundancy,'' \emph{IEEE
  Transactions on Pattern Analysis and Machine Intelligence}, vol.~27, no.~8,
  pp. 1226--1238, 2005.

\bibitem{tran2014overview}
B.~Tran, B.~Xue, and M.~Zhang, ``Overview of particle swarm optimisation for
  feature selection in classification,'' in \emph{Simulated Evolution and
  Learning: 10th International Conference, SEAL 2014, Dunedin, New Zealand,
  December 15-18, 2014. Proceedings 10}.\hskip 1em plus 0.5em minus 0.4em\relax
  Springer, 2014, pp. 605--617.

\bibitem{Deak2022-dv}
B.~Deak, T.~Eggert, A.~Mayr, A.~Stankewitz, F.~Filippopulos, P.~Jahn,
  V.~Witkovsky, A.~Straube, and E.~Schulz, ``Intrinsic network activity
  reflects the fluctuating experience of tonic pain,'' \emph{Cereb. Cortex},
  Jan. 2022.

\bibitem{kohoutova2022individual}
L.~Kohoutov{\'a}, L.~Y. Atlas, C.~B{\"u}chel, J.~T. Buhle, S.~Geuter, M.~Jepma,
  L.~Koban, A.~Krishnan, D.~H. Lee, S.~Lee \emph{et~al.}, ``Individual
  variability in brain representations of pain,'' \emph{Nature Neuroscience},
  pp. 1--11, 2022.

\bibitem{Simis2022}
M.~Simis, M.~Imamura, K.~Pacheco-Barrios, A.~Marduy, P.~S. de~Melo, A.~J.
  Mendes, P.~E.~P. Teixeira, L.~Battistella, and F.~Fregni, ``{EEG} theta and
  beta bands as brain oscillations for different knee osteoarthritis phenotypes
  according to disease severity,'' \emph{Scientific Reports}, vol.~12, no.~1,
  Jan. 2022.

\bibitem{Chai2017}
X.~Chai, Q.~Wang, Y.~Zhao, Y.~Li, D.~Liu, X.~Liu, and O.~Bai,
  ``\BIBforeignlanguage{en}{A fast, efficient domain adaptation technique for
  cross-domain {electroencephalography(EEG)-based} emotion recognition},''
  \emph{\BIBforeignlanguage{en}{Sensors (Basel)}}, vol.~17, no.~5, May 2017.

\bibitem{Hou:22}
J.~Hou, X.~Ding, J.~D. Deng, and S.~Cranefield, ``Deep adversarial transition
  learning using cross-grafted generative stacks,'' \emph{Neural Networks},
  vol. 149, pp. 172--183, 2022.

\bibitem{Wang:23}
Q.~Wang, F.~Meng, and T.~P. Breckon, ``Data augmentation with norm-ae and
  selective pseudo-labelling for unsupervised domain adaptation,'' \emph{Neural
  Networks}, vol. 161, pp. 614--625, 2023.

\end{thebibliography}
\balance
\end{document}